\documentclass{article} 
\usepackage{arxiv}

\usepackage[utf8]{inputenc} 
\usepackage[T1]{fontenc}  
\usepackage{hyperref}    
\usepackage{url}      
\usepackage{booktabs}    
\usepackage{amsfonts}    
\usepackage{nicefrac}    
\usepackage{microtype}   
\usepackage{lipsum}		
\usepackage{graphicx}
\usepackage{natbib}
\usepackage{doi}

\usepackage{bbm}
\usepackage[bb=boondox]{mathalfa}
\usepackage{mathtools}

\usepackage{times}

\newcommand{\x}{\mathbf{x}}
\newcommand{\h}{\mathbf{h}}
\newcommand{\W}{\mathbf{W}}

\newtheorem{theorem}{Theorem}[section]

\newtheorem{lemma}[theorem]{Lemma}

\title{Analyzing Finite Neural Networks: Can We Trust Neural Tangent Kernel Theory?}

\author{ Mariia Seleznova \\
    LMU Munich\\
    \texttt{seleznova@math.lmu.de}\\
	\And
	  Gitta Kutyniok\\
	  LMU Munich\\
    \texttt{kutyniok@math.lmu.de}\\
}

\hypersetup{
pdftitle={Analyzing Finite Neural Networks: Can We Trust Neural Tangent Kernel Theory?},
pdfauthor={Mariia Seleznova, Gitta Kutyniok},
pdfkeywords={Deep Neural Networks, Neural Tangent Kernel},
}
\newcommand\nnfootnote[1]{%
  \begin{NoHyper}
  \renewcommand\thefootnote{}\footnote{#1}%
  \addtocounter{footnote}{-1}%
  \end{NoHyper}
}

\begin{document}
\maketitle
\nnfootnote{*Accepted for publication in Proceedings of 2nd Annual Conference on Mathematical and Scientific Machine Learning, Proceedings of Machine Learning Research (PMLR) vol 145, 2021.}
\begin{abstract}
 Neural Tangent Kernel (NTK) theory is widely used to study the dynamics of infinitely-wide deep neural networks (DNNs) under gradient descent. But do the results for infinitely-wide networks give us hints about the behavior of real finite-width ones? In this paper, we study empirically when NTK theory is valid in practice for fully-connected ReLU and sigmoid DNNs. We find out that whether a network is in the NTK regime depends on the hyperparameters of random initialization and the network's depth. In particular, NTK theory does not explain the behavior of sufficiently deep networks initialized so that their gradients explode as they propagate through {the} network's layers: the kernel is random at initialization and changes significantly during training in this case, contrary to NTK theory. On the other hand, in the case of vanishing gradients{,} DNNs are in the the NTK regime but become untrainable rapidly with depth. We also describe a framework to study generalization properties of DNNs, in particular {the} variance of network's output function, by means of NTK theory and discuss its limits. 
\end{abstract}


\section{Introduction}
Deep neural networks (DNNs) have gained a lot of popularity in the last decades due to their success in a variety of domains, such as image classification \citep{krizhevsky2012imagenet}, speech recognition \citep{hannun2014deep}, playing games \citep{mnih2013playing}, etc. Consequently, there has been a tremendous interest in {the} theoretical properties of DNNs: expressivity \citep{montufar2014number}, optimization \citep{goodfellow2014qualitatively} and generalization \citep{hardt2016train}. However, many aspects of DNNs, in particular their surprising generalization properties, still remain unclear to the community \citep{zhang2016understanding}. 
  
To study theoretical properties of DNNs, numerous recent papers have considered them in the infinite-width limit. In particular, there is a line of research which shows that untrained fully-connected networks of depth $L$ and widths $M_1,\dots,M_L$ with weights and biases initialized randomly as
\begin{equation}\label{eq:rand_init}
\W^l_{ij} \sim \mathcal{N}(0, \sigma_w^2/M_l), \mathbf{b}^l_i \sim \mathcal{N}(0,\sigma_b^2)
\end{equation}
behave as Gaussian processes (GP) in the infinite-width limit (for any $l\in[1,L], M_l \to \infty $) \citep{lee2017deep, matthews2018gaussian, novak2018bayesian}. These GPs are then fully described by a so-called Neural Network Gaussian Process (NNGP) kernel, and a number of publications have studied properties of this kernel depending on the network's depth and initialization hyperparameters
\citep{poole2016exponential,schoenholz2016deep}. These works 
   developed a \textit{mean field} theory formalism for NNs and identified that there exist two situations -- depending on hyperparameters $(\sigma_w^2,\sigma_b^2)$ -- in which signal propagation through the network differs substantially: \textit{ordered} and \textit{chaotic} phases, which correspond to vanishing and exploding gradients. However, these results only concern untrained randomly initialized networks.

  There have also been recent successes in {the} theory of trained infinitely wide DNNs. 
  In particular, it has been shown that {the} evolution of NN's output during gradient flow training can be captured by a so-called Neural Tangent Kernel (NTK) $\Theta^t$ \citep{jacot2018neural,arora2019exact,yang2020tensor}: 
  \begin{equation}\label{eq:grad_flow}
  \begin{split}
  \dfrac{df^t(x)}{dt} = -\dfrac{1}{S}\sum_{s=1,\dots S}\Theta^t(x,x_s) \cdot [f^t(x_s)- y_s],\\
  \Theta^t(x_i,x_j) = \nabla_{w}f^t(x_i)^T\nabla_{w}f^t(x_j),\quad w = \{\W^l,\mathbf{b}^l\}_{l=1,\dots L},
  \end{split}
  \end{equation}
  where $f^t(x)$ is the network's output on $x$ at time $t$ and $D = \{(x_s,y_s)\}_{s=1,\dots S}$ is the training set. In general, {the} NTK changes during training time $t$ and the dynamics in (\ref{eq:grad_flow}) is complex. However, as layers' widths tend to infinity with fixed depth, it can be shown that {the} NTK stays constant during training and equal to its initial value: 
  \begin{equation}\label{eq:the NTK_const_train}
    \Theta^t(x_i,x_j) = \Theta^0(x_i,x_j).
  \end{equation}
  Moreover, {the} NTK at initialization converges to a deterministic kernel $\Theta^{*}$ in the same limit: 
   \begin{equation}\label{eq:the NTK_determ_init}
   \Theta^0(x_i,x_j)\xrightarrow[M_l\to\infty]{} \Theta^{*}(x_i,x_j).
   \end{equation}
These two results allow to dramatically simplify the analysis of DNNs behavior, as the dynamics in (\ref{eq:grad_flow}) becomes identical to kernel regression and the ODE has a closed-formed solution.
  
  However, some recent papers argue that the success of DNNs cannot be explained by their behavior in the infinite-width limit \citep{chizat2019lazy,hanin2019finite}. One justification for this view is that no feature learning occurs when (\ref{eq:the NTK_const_train}) and (\ref{eq:the NTK_determ_init}) hold, as {the} NTK stays constant during training and depends only on the parameters at initialization. Moreover, {the} NTK becomes completely data-independent in the infinite-depth limit, which suggests poor generalization performance \citep{xiao2019disentangling}. That is why, to study properties of real DNNs, it is important to understand when and if NTK theory can be applied to finite-width NNs.

  {\subsection{Contribution}}
  Our aim in this work is to understand when the inferences of NTK theory (\ref{eq:the NTK_const_train}) and (\ref{eq:the NTK_determ_init}) hold for real NNs depending on hyperparameters $(\sigma_w^2, \sigma_b^2,L,M)$ and what this implies for the existing theoretical results about DNNs based on NTK theory. The contributions of our work are as follows:
  \vspace{0mm}
  \begin{itemize}
  \item \textbf{NTK variance at initialization.} We study empirically when {the} NTK is approximately deterministic at initialization for finite-width fully-connected ReLU and \texttt{tanh} networks with different hyperparameters $(\sigma_w^2, \sigma_b^2, L, M)$. Our results suggest that, depending on the initialization hyperparameters ($\sigma_w^2,\sigma_b^2)$, there is a phase in the hyperparameter space where {the} NTK is close to deterministic for any depth $L$, so (\ref{eq:the NTK_determ_init}) holds. However, there is also a phase where {the} NTK variance grows with $L/M$, so (\ref{eq:the NTK_determ_init}) does not hold for deep networks. Following the terminology from \citet{poole2016exponential}, we will call these phases \textit{ordered} and \textit{chaotic}, respectively. 
  
  \item \textbf{NTK change during training.} We also empirically study changes in {the} NTK matrix during gradient descent training for ReLU and \texttt{tanh} networks. Our results show that{,} in the ordered phase, the relative change in {the} NTK matrix norm caused by training is small and does not increase with $L$, so (\ref{eq:the NTK_const_train}) holds. However, in the chaotic phase {the NTK matrix change} during training is large and grows with depth $L$. This implies that (\ref{eq:the NTK_const_train}) does not hold, i.e. DNNs initialized in the chaotic phase do not behave as NTK theory suggests. 
  
  \item \textbf{NTK theory approach for generalization.} Some recent publications analyze properties of {the} NTK and draw conclusions about DNNs' generalization thereof \citep{xiao2019disentangling, geiger2020scaling}. Other authors argue that the behavior of networks in {the} NTK regime is trivial and does not yield good generalization properties, {that are however observed for DNNs in practice} \citep{chizat2019lazy}. We show how to compute data-independent variance of {the} network's output when it evolves according to NTK theory. However, given our empirical results {for} when NTK theory is applicable, we discover that these findings do not explain the behavior of finite-width networks in most of the hyperparameters space $(\sigma_{{w}}^2,\sigma_b^2,L,M)$.
\end{itemize}
  
   
  
{\subsection{Related work}}
 {This work adds to the line of research that studies the correspondence between finite- and infinite-width DNNs. In particular, the difference between theoretical (infinite-width) and empirical (finite-width) NTK. In this section, we survey the prior results in this direction and position our contribution within them.
 
 A number of papers have studied the convergence of the empirical NTK at initialization to the theoretical NTK. The first fundamental result of NTK theory is that the NTK converges to a deterministic limit as $M$ goes to infinity \citep{jacot2018neural}. The following work proved a non-asymptotic bound on minimal $M$ required to guarantee this convergence in case of ReLU networks \citep{arora2019exact}. This bound on $M$ depends on the depth as $O(L^6log(L))$, therefore $L/M$ is always small for deep networks when the bound holds. Then, a recent theoretical work improved this result in a special case of ReLU networks with initialization ($\sigma_w=2,\sigma_b=0$) by showing the precise exponential dependence of the NTK variance at initialization on $L/M$ \citep{hanin2019finite}. That is, (\ref{eq:the NTK_determ_init}) does not hold for such networks when $L/M$ is bounded away from zero. However, the proofs given in the paper are not immediately generalizable for different activation functions and different initialization parameters. Thus, there is still no solid understanding of the NTK randomness depending on the choice of a network. Therefore, in Section \ref{section:the NTK_init_var}, we empirically study the randomness of the NTK at initialization for ReLU and \texttt{tanh} networks with a variety of hyperparameters ($M,L,\sigma_w,\sigma_b$) and observe the precise dependence on 1) the position of initialization ($\sigma_w,\sigma_b$) in either ordered or chaotic phase, 2) depth-to-width ratio $L/M$ in the chaotic phase.

Changes of the NTK matrix during gradient descent training have also been analyzed in the literature mostly as a function of $M$. In particular, it has been proven \citep{huang2020dynamics} and shown experimentally \citep{lee2019wide} that the change of the NTK matrix during gradient descent training is bounded by $O(1/M)$ when the depth $L$ is fixed. For ReLU networks with initialization ($\sigma_w=2,\sigma_b=0$) it has also been proven that the change of the NTK in a gradient descent step depends exponentially on $L/M$ \citep{hanin2019finite}. We add to these results in Section \ref{section:the NTK_change_train} by investigating the NTK changes during training for two activation functions and hyperparameters $(\sigma_w,\sigma_b,L)$. 

A different line of research has also studied the theoretical (infinite-width) NTK as a function of depth and initialization parameters \citep{xiao2019disentangling,hayou2019mean}. These contributions found that the spectrum of infinite-width NTK behaves differently in ordered and chaotic phases. The authors also showed that the infinite-\textit{depth} limit of the theoretical NTK (when first the limit $M\to\infty$ is taken with fixed $L$ and then $L\to\infty$) yields trivial performance and cannot explain properties of finite DNNs. These papers showed that both in ordered and chaotic phases the NTK approaches its trivial limit exponentially in $L$, and only in the border between phases (EOC) this convergence is sub-exponential. However, the setting of these contributions requires $L/M$ values to be small, therefore they do not explain how the randomness of NTK and its changes during training impact the results. Our work shows that in the chaotic phase and at the EOC the NTK does not behave as its theoretical limit when $L/M$ is bounded away from zero, therefore we cannot draw conclusions about such DNNs based on the theoretical NTK. 

In generalization research, the recent trend is double descent -- the phenomenon that highly overparametrized models, including DNNs, tend to generalize surprisingly well \citep{belkin2018reconciling,nakkiran2019deep,belkin2019two,hastie2019surprises}. The recent developments in the theory of double descent showed that overparametrized linear models reach low generalization error because, counterintuitively, their variance decreases when the number of parameters increases beyond the number of samples \citep{hastie2019surprises}. However, there is still no double descent theory for DNNs, which are significantly more theoretically complex than linear models. In Section \ref{section:generalization}, we studied the variance of DNNs' output with the simplifications of NTK theory, which can be seen as the first step into this direction. }
\vspace{5mm}

\section{Mean field approach for wide neural networks}\label{section:mean_field_theory}

A number of recent papers used {the} \textit{mean field} formalism to study forward- and backpropagation of signal through randomly initialized DNNs \citep{poole2016exponential,schoenholz2016deep,karakida2018universal,yang2017mean}. We first describe this approach and show how ordered and chaotic phases, which correspond to vanishing and exploding gradients, arise from it.

Suppose there is a fully-connected feed-forward neural network initialized randomly as in (\ref{eq:rand_init}) with hidden layers' widths $M_1,\dots M_L$. 
Forward propagation through the network is given by
\begin{equation*}
\begin{split}
  \x^l(x_s) = \phi(\h^l(x_s)),\quad \h^l(x_s) = \W^l\x^{l-1}(x_s) + \mathbf{b}^l, \quad l=1,\dots L, \\
  \x^{0}(x_s)=x_s, \quad s = 1,\dots S,
\end{split}
\end{equation*}
where $\phi$ is the activation function, $\x^l$ are activations, $\h^l$ are pre-activations in each layer $l$, and $D = (X,Y) = \{(x_s,y_s)\}_{s=1,\dots S}$ is a dataset. 

Consider variances $q^l(x_s) \coloneqq \mathbb{E}[(\h_i^l(x_s))^2]$ of the pre-activations in each layer for a given input vector $x_s$. The mean field theory approach assumes that $\h_i^l(x_s), \ i=1,\dots M_l$ are i.i.d Gaussian, so by central limit theorem in the limit of $M \to \infty$, the variance can be seen as a sum over different neurons in the same layer $q^l(x_s) = \frac{1}{M_l}\sum_{i=1}^{M_l} (\h_i^l(x_s))^2$. Then it can be computed through a recursive relation:

\begin{equation}\label{eq:q_l}
  q^l(x_s) = \sigma_w^2 \int Dz \cdot \phi(\sqrt{q^{l-1}(x_s)}z)^2 + \sigma_b^2,
\end{equation}
where the average over numerous neurons in layer $l-1$ is replaced by an integral over a Gaussian distribution $Dz = \frac{dz}{\sqrt{2\pi}}e^{-z^2/2}$. Then the variance of activations $\hat{q}^l(x_s) \coloneqq \mathbb{E}[(\x_i^l(x_s))^2]$ is given by

\begin{equation}\label{eq:hat_q_l}
  \hat{q}^l(x_s) = \int Dz \cdot \phi(\sqrt{q^{l}(x_s)}z)^2.
\end{equation}
In the same fashion, \citet{poole2016exponential} derive a recursive map for {the} correlation between pre-activations {of two different inputs} and {the correlation between} activations of two different inputs, denoted correspondigly $q^l(x_s,x_r) \coloneqq \mathbb{E}[\h_i^l(x_s)\h_i^l(x_r)]$ and $\hat{q}^l(x_s,x_r)] \coloneqq \mathbb{E}[\x_i^l(x_s)\x_i^l(x_r)]$:

\begin{equation}\label{eq:q_sr_l}
\begin{split}
  q^l_{sr}(x_s,x_r) = \sigma_w^2 \int Dz_1 Dz_2 \cdot \phi(u_1)\phi(u_2) + \sigma_b^2,\\
  \hat{q}_{sr}^{l-1}(x_s,x_r) = \int Dz_1 Dz_2 \cdot \phi(u_1)\phi(u_2),\\
  u_1 = \sqrt{q^{l-1}(x_s)}z_1, \quad u_2 = \sqrt{q^{l-1}(x_r)}[c_{sr}^{l-1}z_1 + \sqrt{1-(c_{sr}^{l-1})^2}z_2],\\
  c_{sr}^{l-1} = \dfrac{q^{l-1}(x_s,x_r)}{\sqrt{q^{l-1}(x_s)q^{l-1}(x_r)}}.
\end{split}
\end{equation}

The gradients of the network are given by the backpropagation chain:
\begin{equation*}
\begin{split}
  \dfrac{\partial f}{\partial \W_{ij}^l} = \delta_i^l\phi(\h^{l-1}_j),\quad \dfrac{\partial f}{\partial \mathbf{b}_{i}^l} = \delta_i^l, \\
  \delta_i^l = \dfrac{\partial f}{\partial \h_i^l} = \phi^{'}(\h_i^l)\sum_j \delta_j^{l+1}\W_{ji}^{l+1},
\end{split}
\end{equation*}
where we omitted the dependence on input $x_s$ for simplicity. With an additional assumption that weights in forward- and backpropagation are drawn independently, i.e. $\phi(\h^l_j)$ and $\delta_i^l$ are independent, \citet{schoenholz2016deep} derived a recursive relation for the variance of the backpropagated errors $p^l(x_s) \coloneqq \mathbb{E}[\sum_i(\delta_i^l(x_s))^2]$:

\begin{equation}\label{eq:p_l}
  p^l(x_s) = \sigma_w^2 p^{l+1}(x_s) \dfrac{M_{l+1}}{M_{l+2}} \int Dz [\phi^{'}(\sqrt{q^{l}(x_s)}z)]^2.
\end{equation}
And for the corresponding correlation between backpropagated errors of two different input vectors $p^l_{sr}(x_s,x_r) \coloneqq \mathbb{E}[\sum_i(\delta_i^l(x_s)\delta_i^l(x_r))]$:
\begin{equation}\label{eq:p_sr_l}
\begin{split}
  p^l_{sr}(x_s,x_r) = \sigma_w^2 p^{l+1}_{sr}(x_s,x_r) \dfrac{M_{l+1}}{M_{l+2}} \int Dz_1 Dz_2 \cdot \phi^{'}(u_1)\phi^{'}(u_2),\\
  u_1 = \sqrt{q^{l}(x_s)}z_1, \quad u_2 = \sqrt{q^{l}(x_r)}[c_{sr}^{l}z_1 + \sqrt{1-(c_{sr}^{l})^2}z_2],\\
  c_{sr}^{l} = \dfrac{q^{l}_{sr}(x_s,x_r)}{\sqrt{q^{l}(x_s)q^{l}(x_r)}}.
\end{split}
\end{equation}
Note that for certain activation functions, e.g. ReLU and \texttt{erf}, the integrals in (\ref{eq:q_l}), (\ref{eq:hat_q_l}), (\ref{eq:q_sr_l}), (\ref{eq:p_l}) and (\ref{eq:p_sr_l}) can be taken analytically. One can refer to Appendix \ref{appendix:integrals} for these analytical expressions.

We can now introduce, following the notation from \citet{poole2016exponential} and \citet{schoenholz2016deep}, a quantity that controls the backpropagation of variance $p^l(x_s)$:
\begin{equation*}
\begin{split}
  \chi_1^l = \sigma_w^2 \int Dz [\phi^{'}(\sqrt{q^{l}}z)]^2, \\
  p^l = p^{l+1} \cdot \chi_1^l,
\end{split}
\end{equation*}
where we assumed that the network's width is constant, i.e. $M_{l+1}/M_{l+2} = 1$. Then $\chi_1$ also controls the propagation of the gradients at initialization:
\begin{equation*}
\begin{split}
  \mathbb{E}[(\dfrac{\partial f^0(x_s)}{\partial \W_{ij}^l})^2 ] = \mathbb{E}[(\delta_i^l)^2]\mathbb{E}[(\phi(\h_j^{l-1}))^2] \propto p^l(x_s).
\end{split}
\end{equation*}
In particular, when the initialization parameters are such that $\chi_1^l < 1$ in all the layers, the gradients vanish, and when $\chi_1^l > 1$ the gradients explode. These two situations are referred to as \textit{ordered} and \textit{chaotic} phases correspondingly, and the border between these phases defined by $\chi_1^l = 1 $ is called \textit{edge of chaos} (EOC) initialization. Several authors suggest that networks should be initialized near EOC to allow deeper signal propagation \citep{hayou2018selection,schoenholz2016deep}.

In the next two sections of the paper, we test empirically how different parameters of random initialization $(\sigma_w^2,\sigma_b^2)$, as well as network's architecture $(M,L)$, impact the behavior of {the} empirical NTK $\Theta^t$. Our observation is that for finite-width networks chaotic and ordered phases give rise to very different behavior of {the} empirical NTK as compared to {the} theoretical NTK, which has not been considered in the community before to the best of our knowledge.

\section{NTK variance at initialization}\label{section:the NTK_init_var}
\begin{figure}
\begin{center}
\includegraphics[width=1.\linewidth]{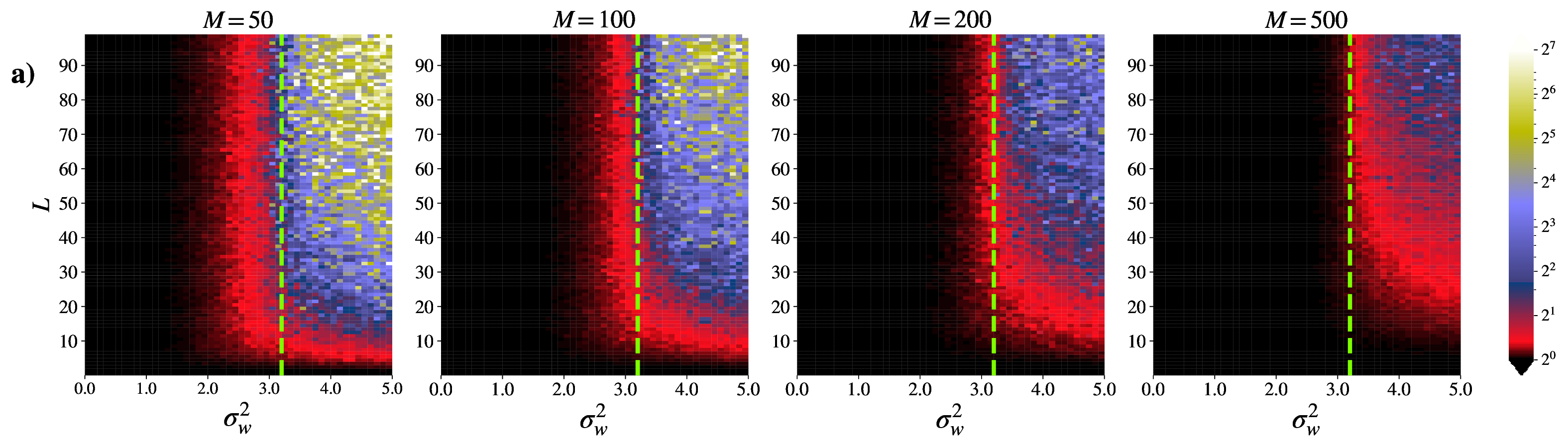}
\includegraphics[width=1.\linewidth]{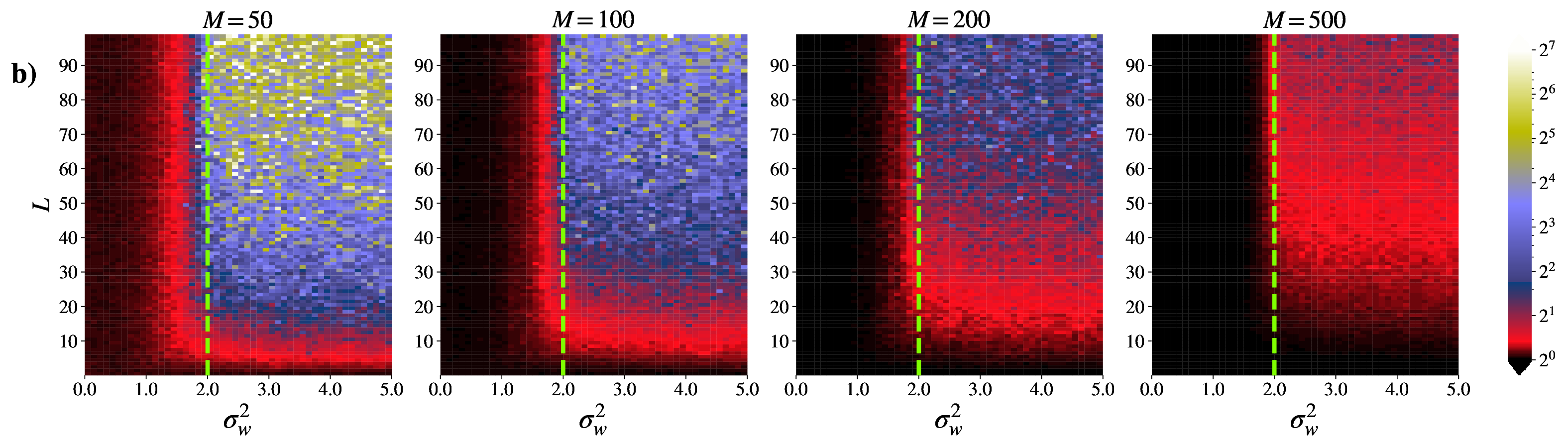}
\end{center}
  \caption{{Ratio} $\dfrac{\mathbb{E}[\Theta^{0}(x,x)^2]}{\mathbb{E}^2[\Theta^{0}(x,x)]}$ for fully-connected a) \texttt{tanh}, b) ReLU networks of constant widths $M=50,100,200,{500}$, in all the experiments $\sigma_b^2=1$. The expected values for each set of parameters are calculated by sampling 200 random initializations of the network. {The} NTK is computed using TensorFlow automatic differentiation. The dashed line shows the theoretical border between ordered and chaotic phases ($\chi^l_1=1$) for the given hyperparameters. In the black zone, the ratio is close to one, i.e. {the} NTK at initialization $\Theta^{0}$ has low variance and can be considered a deterministic variable. In the red zone, the NTK standard deviation is comparable with its mean. In the blue zone, the NTK standard deviation is greater than its mean, so {the} NTK is not deterministic and cannot be replaced by its mean.}
\label{fig_relu_Kxx_var}
\end{figure}
\begin{figure}
\begin{center}
\includegraphics[width=1.\linewidth]{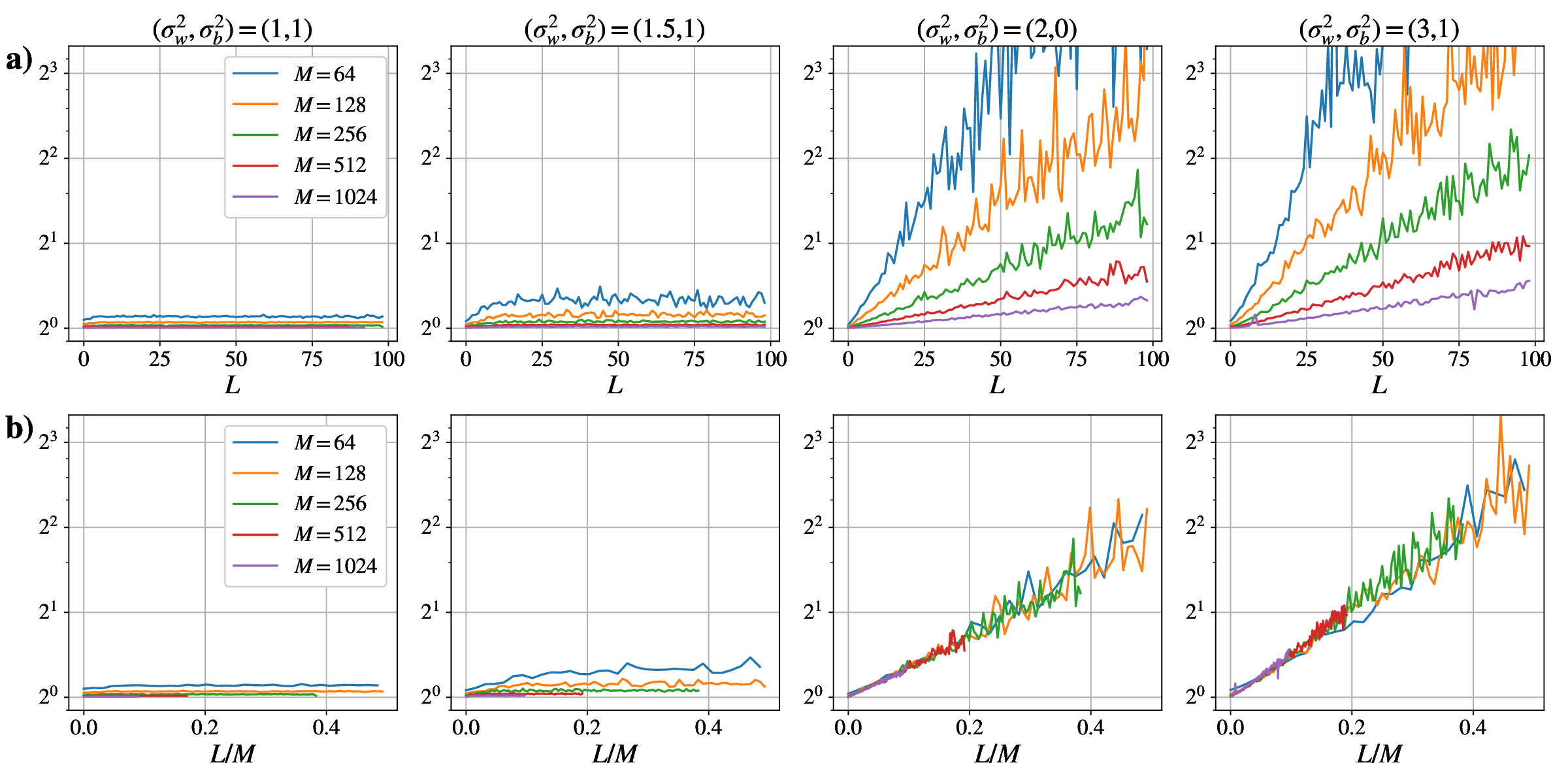}
\end{center}
  \caption{Dependence of {ratio} $\dfrac{\mathbb{E}[\Theta^{0}(x,x)^2]}{\mathbb{E}^2[\Theta^{0}(x,x)]}$ on $L/M$ with different initialization parameters and width values {for ReLU networks}. Both rows show the same curves plotted against a) depth $L$, b) ratio $L/M$. The expectations are computed by sampling 200 random initializations of the network.}
\label{fig:lm_ratio}
\end{figure}

First we aim to verify empirically when the theoretical result (\ref{eq:the NTK_determ_init}) that {the} NTK is deterministic at initialization in the infinite-width limit holds for finite-width NNs. Following \citet{hanin2019finite}, we computed the ratio $\mathbb{E}[\Theta^{0}(x,x)^2]/{\mathbb{E}^2[\Theta^{0}(x,x)]} \in [1,\infty)$ to study the distribution of {the} NTK. When {the} NTK at initialization is close to deterministic, its distribution is similar to a delta function around its mean and the value of the ratio is close to one. On the other hand, when this ratio is bounded away from one, the NTK's variance is comparable to its mean value and therefore cannot be disregarded.

One can see the results of our experiments for fully-connected ReLU and \texttt{tanh} networks with constant width $M$ in Figure \ref{fig_relu_Kxx_var}. We observe that when $\sigma_w^2$ is small enough (ordered phase), {the} NTK variance is small and does not increase with depth $L$, implying that (\ref{eq:the NTK_determ_init}) holds for any depth and NTK theory can be used to study NNs initialized in this way. However, for large $\sigma_w^2$ (chaotic phase) the variance grows significantly with $L$, hence for very deep networks in this phase (\ref{eq:the NTK_determ_init}) does not hold. At the EOC, the variance of {the} NTK is a fraction of its mean even for very deep networks, so NTK theory can approximate the average behavior of networks initialized near EOC, but the random effects may still be significant. {One can also see that as $M$ grows, the vertical red region gets narrower, i.e. the transition becomes sharper. This is consistent with the fact that the theoretical border between vanishing and exploding gradients is sharp and computed in mean field theory (Section \ref{section:mean_field_theory}) by taking the limit $M\to\infty$.} These results are similar for ReLU and \texttt{tanh} networks, taking into account that the theoretical boundary between phases --- given by $\chi_1^l=1$ and indicated by the dashed line in the figures --- is located at larger $\sigma_w^2$ values for sigmoid networks. One also observes that {the} NTK variance is small for sufficiently shallow NNs with any $\sigma_w^2$ value. Such shallow networks were mostly considered in recent empirical studies on behavior of wide NNs under gradient descent \citep{lee2019wide}. It is thus important to note, that such empirical results may be invalid for much deeper networks, depending on the initialization parameters.

Moreover, when depth $L$ is fixed and width $M$ increases, the the NTK variance decreases {in the chaotic phase}, which supports the hypothesis that the variance depends on the ratio $L/M$. To examine this dependence {on} $L/M$ in more detail, we present Figure \ref{fig:lm_ratio}. It shows the {ratio} $\mathbb{E}[\Theta^{0}(x,x)^2]/{\mathbb{E}^2[\Theta^{0}(x,x)]}$ for a wider range of $M$ values for {four} different initialization parameters sets: $(\sigma_w^2,\sigma_b^2) \in [(1,1),{(1.5,1)},(2,0), (3,1)]$. Each curve is plotted against both $L$ and $L/M$. We notice that in the ordered phase ($\sigma_w^2=1$ {and $\sigma_w^2=1.5$}) the ratio is close to 1, does not grow with $L/M$ and decreases with $M$. {In this phase, the NTK converges to its deterministic limit with increasing $M$ regardless of the $L$ value, which is the expected behaviour within NTK theory.} However, in the chaotic phase ($\sigma_w^2=3$) the ratio grows exponentially as a function of $L/M$. {This observation gives a precise scaling for minimal $M$ values required to assume that the NTK of a network with a given depth $L$ is deterministic at initialization, which improves the previous asymptotic result in \citet{jacot2018neural} and the bound on required $M$ in \citet{arora2019exact}.} In case of ReLU networks and initialization $(\sigma_w^2,\sigma_b^2)=(2,0)$, \citet{hanin2019finite} theoretically showed that the $\mathbb{E}[\Theta^{0}(x,x)^2]/{\mathbb{E}^2[\Theta^{0}(x,x)]}$ ratio is {indeed} exponential in $L/M$, but their analysis is not trivially generalizable for different activation functions and initialization parameters. {Our experiments confirm these findings in the special case but also show that changing initialization parameters impacts the behaviour of the the NTK variance significantly.}

We also checked if the value of $\sigma_b^2$ impacts {the NTK variance behavior} at initialization significantly. {In Appendix \ref{appendix:the NTK_train}, we provide figures showing the NTK variance with different $\sigma_b^2$ values. We observed that lower $\sigma_b^2$ values} yield narrower boundary between the two phases identified in Figure \ref{fig_relu_Kxx_var}, but the general picture stays similar. 

\section{NTK change during training}\label{section:the NTK_change_train}
\begin{figure}
\begin{center}
\includegraphics[width=1.\linewidth]{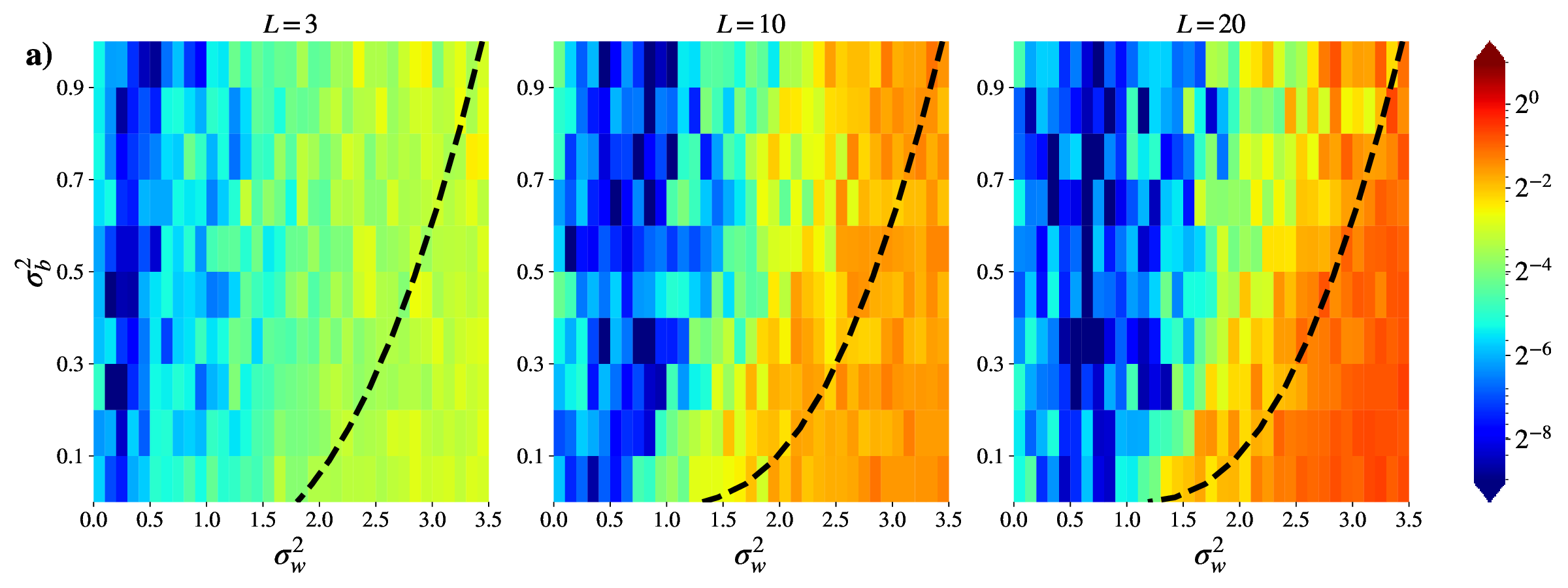}
\includegraphics[width=1.\linewidth]{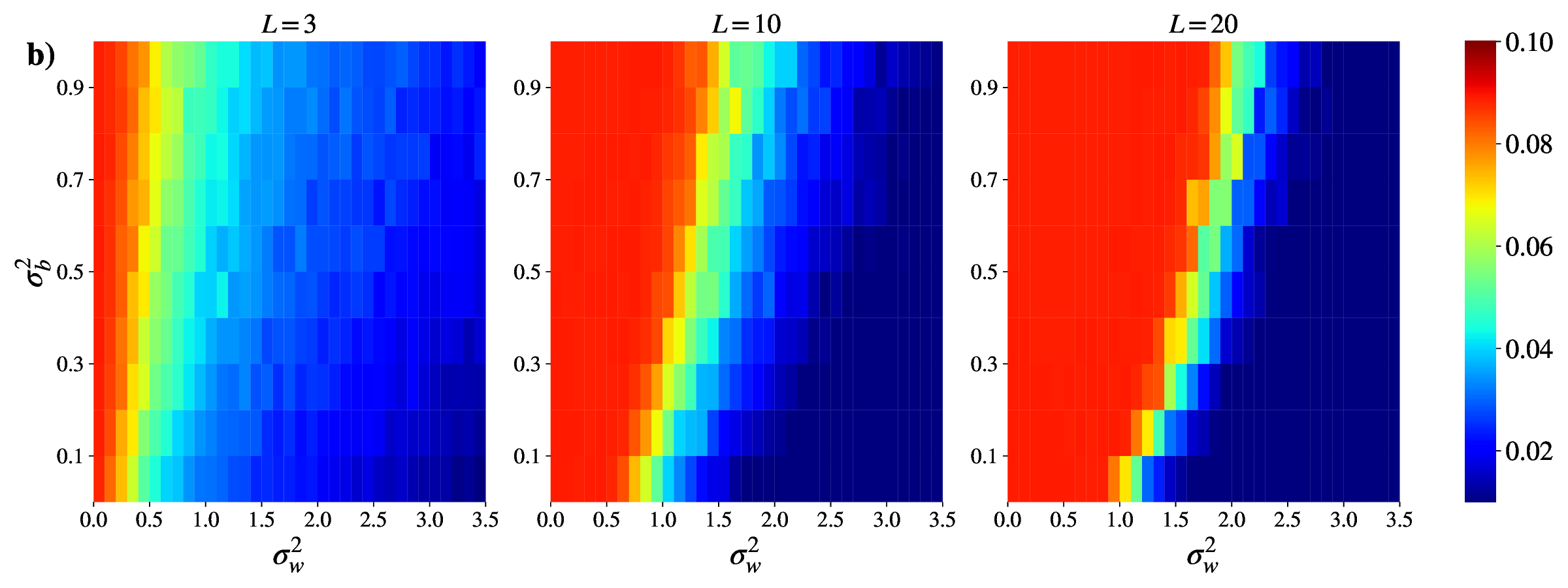}
\end{center}
  \caption{ a) Relative change in {the} NTK norm $\dfrac{\|\Theta^t-\Theta^{0}\|_F}{\|\Theta^{0}\|_F}$ for \texttt{tanh} networks {of width $M=256$} trained by gradient descent {with MSE loss} on a subset of MNIST (128 samples). The dashed line indicates the theoretical border between ordered and chaotic phases {($\chi_1^l=1$)}. {We used early stopping when the loss did not decrease by at least $10^{-7}$ in 100 consecutive steps, otherwise the number of training steps was limited by $10^5$.} The learning rate is constant and equals $10^{-5}$ for all the networks{, which is chosen so that, for all the hyperparameters, it does not exceed the theoretical maximal learning rate for wide networks derived in \citet{karakida2018universal}.} b) Minimal loss value that the network{s} manage{d} to reach in our experiments. Networks in the red area are untrainable with the given learning rate, networks in the blue area are trainable. }
\label{fig:tanh_train_heatmap}
\end{figure}

\begin{figure}
\begin{center}
\includegraphics[width=1.\linewidth]{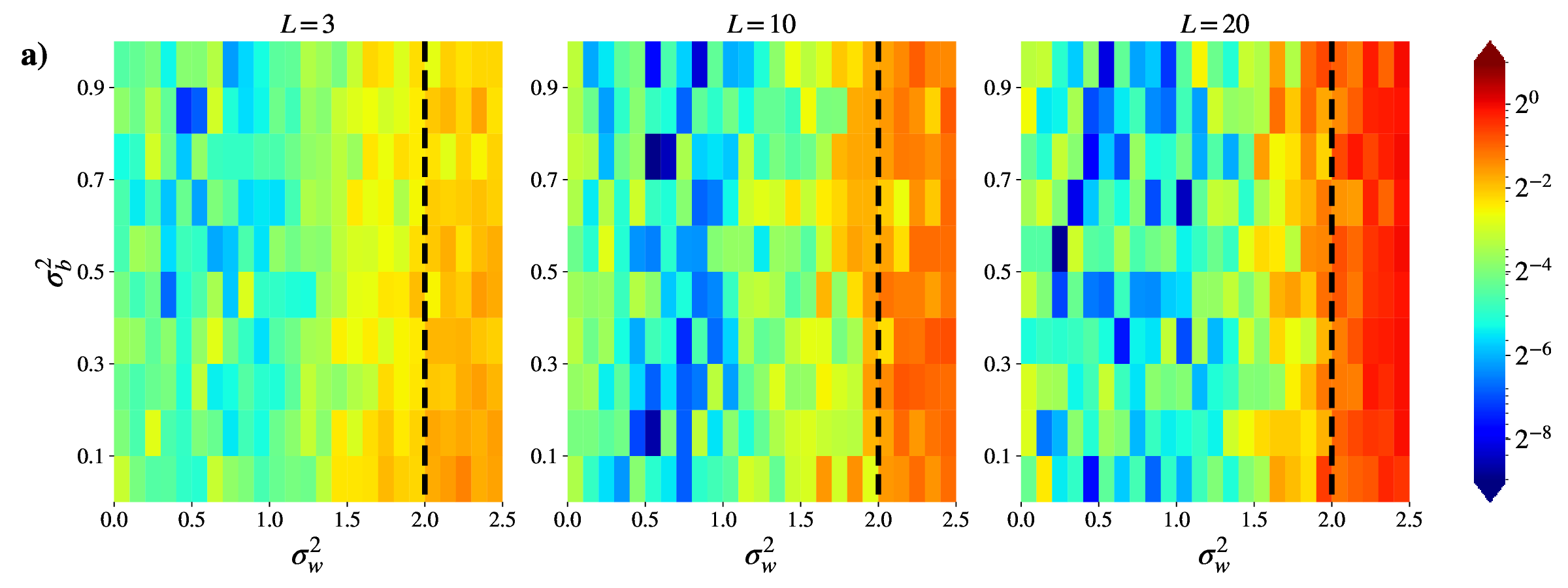}
\includegraphics[width=1.\linewidth]{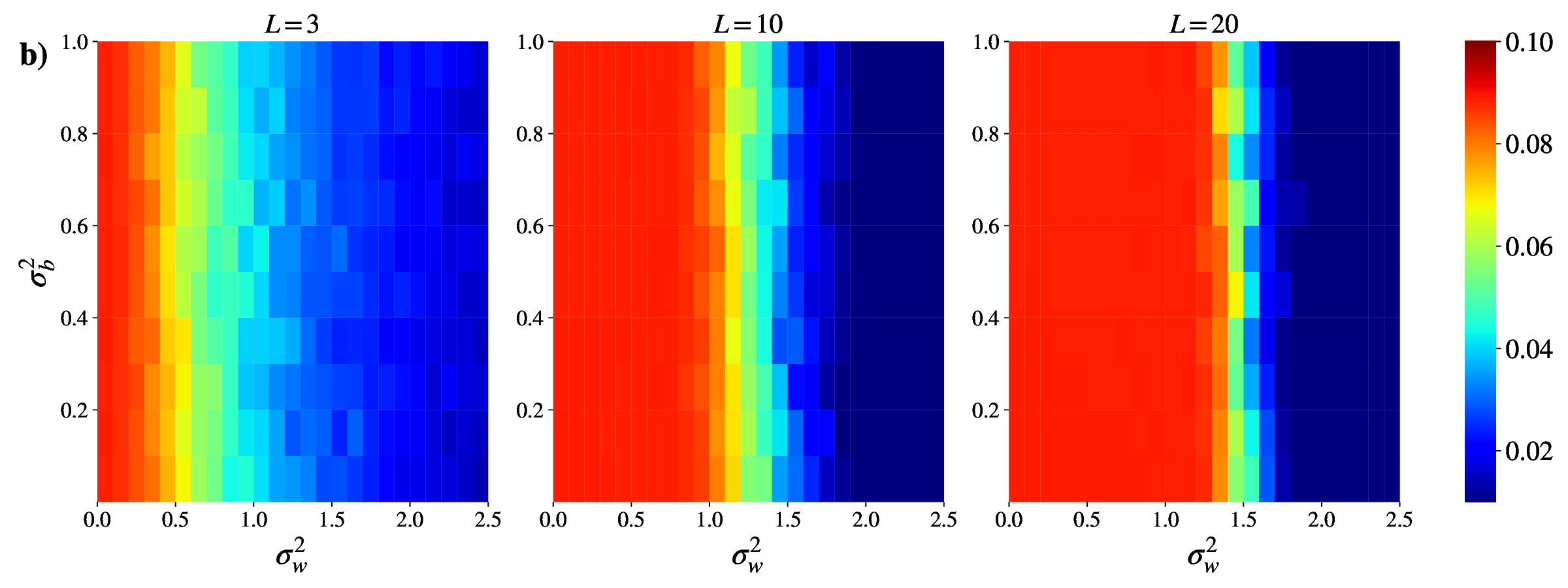}
\end{center}
  \caption{ a) Relative change in {the} NTK norm $\dfrac{\|\Theta^t-\Theta^{0}\|_F}{\|\Theta^{0}\|_F}$ for ReLU networks {of width $M=256$} trained by gradient descent {with MSE loss} on a subset of MNIST (128 samples). The dashed line indicates the theoretical border between ordered and chaotic phases {($\chi_1^l=1$)}. {We used early stopping when the loss did not decrease by at least $10^{-7}$ in 100 consecutive steps, otherwise the number of training steps was limited by $10^5$.} The learning rate is constant and equals $10^{-5}$ for all the networks{, which is chosen so that, for all the hyperparameters, it does not exceed the theoretical maximal learning rate for wide networks derived in \citet{karakida2018universal}.} b) Minimal loss value that the network{s} manage{d} to reach in our experiments. Networks in the red area are untrainable with the given learning rate, networks in the blue area are trainable. }
\label{fig:relu_train_heatmap}
\end{figure}

{In this section we present the numerical experiments that we conducted to check whether the second result of NTK theory (\ref{eq:the NTK_const_train}) holds, i.e.} whether {the} empirical NTK of finite-width ReLU and \texttt{tanh} networks stays approximately constant during training with gradient descent.
We trained networks with a variety of hyperparameters $(\sigma_w^2,\sigma_b^2,L)$ and measured the relative change of NTK's Frobenious norm ${\|\Theta^t-\Theta^{0}\|_F}/{\|\Theta^{0}\|_F}$ that occurs during training.
{The results for \texttt{tanh} and ReLU networks are in Figures \ref{fig:tanh_train_heatmap}a and \ref{fig:relu_train_heatmap}a. In Figures \ref{fig:tanh_train_heatmap}b and \ref{fig:relu_train_heatmap}b, we also plotted the minimal losses that the networks reached in the experiments.}

We draw the following conclusions from the experiments' results:
\begin{itemize}
  \item \textbf{Phase transition for empirical NTK}. For both ReLU and \texttt{tanh} networks, {the} NTK behavior during training changes significantly around the theoretical border between chaotic and ordered phases.
  \item \textbf{Chaotic phase}. In the chaotic phase, the relative change in {the} NTK matrix norm is significant and increases with depth $L$, so one cannot assume that the kernel stays constant during training for deep networks. However, for very shallow networks the NTK at initialization may still be a good approximation for {the} NTK after training. In the previous section we also saw that {the} NTK matrix of shallow networks in the chaotic phase is close to deterministic at initialization, which shows that NTK theory approximates only shallow networks in the chaotic phase.
  \item \textbf{Ordered phase}. In the ordered phase, the relative change in {the} NTK matrix norm is small throughout training for any depth. We saw in the previous section that {the} NTK is also close to deterministic at initialization in this phase. It follows that in the ordered phase finite-width DNNs behave as NTK theory suggests even when depth $L$ is large. 
  \item \textbf{EOC}. There is a region close to the border between phases where the change in {the} NTK norm is larger than in the ordered phase but still remains way below 1 for deep networks. We also saw in the previous section that in this region the standard deviation of {the} NTK is lower than its mean value for deep networks. Thus, NTK theory can approximate behavior of deeper networks in case of EOC initialization in comparison to the chaotic phase, but the effects of randomness and change during training may still play a significant role.
  \item \textbf{Trainability}. Networks become untrainable with depth much faster in the ordered phase than in the chaotic phase. In our experiments, networks in the ordered phase with $L=20$ already mostly cannot reach low training loss values. This is consistent with the results on trainability provided in \citet{xiao2019disentangling}.
\end{itemize}
We thus have discovered two regions in the hyperparameters space $(\sigma_w^2,\sigma_b^2,L,M)$ where both statements of NTK theory (\ref{eq:the NTK_const_train}) and (\ref{eq:the NTK_determ_init}) hold: the ordered phase with any depth $L$ and the chaotic phase where the $L/M$ ratio is low. For other choices of architecture and initialization, our experiments suggest that finite-width networks do not behave according to NTK theory.

{ Note that the networks in Figures \ref{fig:tanh_train_heatmap}a and \ref{fig:relu_train_heatmap}a take different number of training steps to reach their final loss values. Somewhat counterintuitively, we observe that the networks which take more iterations to train show mostly small changes in the NTK matrix norm. To provide more insight about the NTK dynamics during different stages of training, we also include figures that show changes in the NTK matrix norm as a function of the number of training steps, as well as figures with changes of the NTK for different $M$ values, in Appendix \ref{appendix:the NTK_train}. } 

\section{NTK theory approach for generalization}\label{section:generalization}

If {the} NTK stays constant during training (\ref{eq:the NTK_const_train}), then the dynamics in (\ref{eq:grad_flow}) are identical to kernel regression with kernel $\Theta^{0}$. In such dynamics, the output function of a network that is trained until convergence ($t\to\infty$) by gradient flow with MSE loss is given by:
\begin{equation}\label{eq:f_t}
\begin{split}
  f^{t=\infty}(x) = \Theta^{0}(x,X)\Theta^{0}(X)^{-1}Y + f^0(x) - \Theta^{0}(x,X)\Theta^{0}(X)^{-1}f^0(X),
\end{split}
\end{equation}
where $\Theta^0(X)$ is the kernel matrix of all the pairs of inputs in $X = [x_s]_{s=1,\dots S}$, i.e. $\Theta(X) = [\Theta^0(x_s,x_r)]_{s,r=1,\dots S}$, and $\Theta(x,X) = [\Theta^0(x,x_s)]_{s=1,\dots S}$ and $f^0(X)=[f^0(x_s)]^T_{s=1,\dots S}$. One can refer to \citet{arora2019exact} or \citet{lee2019wide} for the derivation of this equation. If {the} NTK is also deterministic at initialization (\ref{eq:the NTK_determ_init}), then the only variables in (\ref{eq:f_t}) that are random with respect to the network's parameters at initialization $w_0$ are $f^0(x)$ and $f^0(X)$, which greatly simplifies the analysis of the generalization properties of $f^{t=\infty}$. 

Let us denote $R(x) \coloneqq \mathbb{E}_{w_0,D}[(f^{t=\infty}(x)-y_{true})^2]$ -- the expected error on an arbitrary test point $x$, given that the initialization is random. Then we can write the bias-variance decomposition as follows:

\begin{equation*}
\begin{split}
   R(x) = Var(f^{t=\infty}(x)) + Bias(f^{t=\infty}(x)),
\end{split}
\end{equation*}
where
\begin{equation*}
\begin{split}
Var(f^{t=\infty}(x)) &= \mathbb{E}_{w_0,D}[(f^{t=\infty}(x)- \mathbb{E}_{w_0,D}[f^{t=\infty}(x)])^2],\\
Bias(f^{t=\infty}(x)) &=\mathbb{E}_{w_0,D}[(\mathbb{E}_{w_0,D}[f^{t=\infty}(x)]-y_{true})^2 ].
\end{split}
\end{equation*}
Then NTK theory allows us to analyze the variance term to characterize the generalization error of the network $\mathbb{E}_{x}[R(x)]$. To do so, first let us show how distributions of the terms in (\ref{eq:f_t}) can be characterized by the mean field theory quantities introduced in Section \ref{section:mean_field_theory}. First of all, the distribution of the network's output at initialization is given directly by the definitions of $q^L$ and $q^L_{sr}$. Hence, the following lemma is immediate.

\begin{lemma}\label{lemma:f0_distr}
The variance of the output function $f^0$ of a randomly initialized network and the covariance of outputs on two different input vectors are given by:
\begin{equation*}
\begin{split}
  \mathbb{E}[(f^0(x))^2] = \mathbb{E}[(\h^L_i(x))^2] = q^{L}(x),
\end{split}
\end{equation*}
\begin{equation*}
  \mathbb{E}[f^0(x_s)f^0(x_r)] = \mathbb{E}[\h^{L}_i(x_s)\h^L_i(x_r)] = q^L_{sr}(x_s,x_r).
\end{equation*}
\end{lemma}
Recall that the NTK is composed of gradients as $\Theta^{0}(x_s,x_r) = \nabla_{w} f^0(x_s)^T \nabla_{w} f^0(x_s)$ and its expected values are therefore proportional to the variances of gradients, considered in Section \ref{section:mean_field_theory}. Then, assuming that the {the} NTK matrix at initialization is deterministic and equal to its expected value, we can express it through quantities $q^l,p^l,q_{sr}^l,p_{sr}^l$ by the following lemma.
\begin{lemma}\label{lemma:Theta}
For a fully-connected network with widths $M_l=\alpha_l M,l=0,\dots L$ (where $M_0$ is the input dimension), deterministic {the} NTK matrix on a sample $X=\{x_s\}_{s=1,\dots S}$ at initialization is given by:
\begin{equation*}
\begin{split}
  \Theta^{*}(X) = \alpha M \bigl(\Lambda + O(1/M) \bigr), \\
\end{split}
\end{equation*}  
\begin{equation*}
\begin{split}
  \Lambda = &\begin{bmatrix}
\kappa_1(x_1) & \kappa_2(x_1,x_2) & \dots & \kappa_2(x_1,x_S) \\
\kappa_2(x_1,x_2) & \kappa_1(x_2) & & \hdots \\
\dots & & & \kappa_2(x_1,x_{S-1})\\
\kappa_2(x_1,x_{S}) & \dots & \kappa_2(x_1,x_{S-1}) & \kappa_1(x_S)&
\end{bmatrix}, \\
\end{split}
\end{equation*}
\begin{equation*}
\begin{split}
\kappa_1(x) = \sum_{l=1}^{L} \dfrac{\alpha_{l-1}}{\alpha}\hat{q}^{l-1}(x) p^l(x), \quad& \kappa_2(x_s,x_r) = \sum_{l=1}^{L} \dfrac{\alpha_{l-1}}{\alpha} \hat{q}^{l-1}_{sr}(x_s,x_r)p_{sr}^l(x_s,x_r),
\end{split}
\end{equation*}
where $\alpha = \sum_{l=1}^{L-1} \alpha_l\alpha_{l-1}$. 
\end{lemma}
We give a proof for this lemma in Appendix \ref{appendix:the NTK}. We note that the same statement is also proven in \citet{karakida2018universal} as a part of Theorem 3.
\begin{figure}
\begin{center}
\includegraphics[width=1.\linewidth]{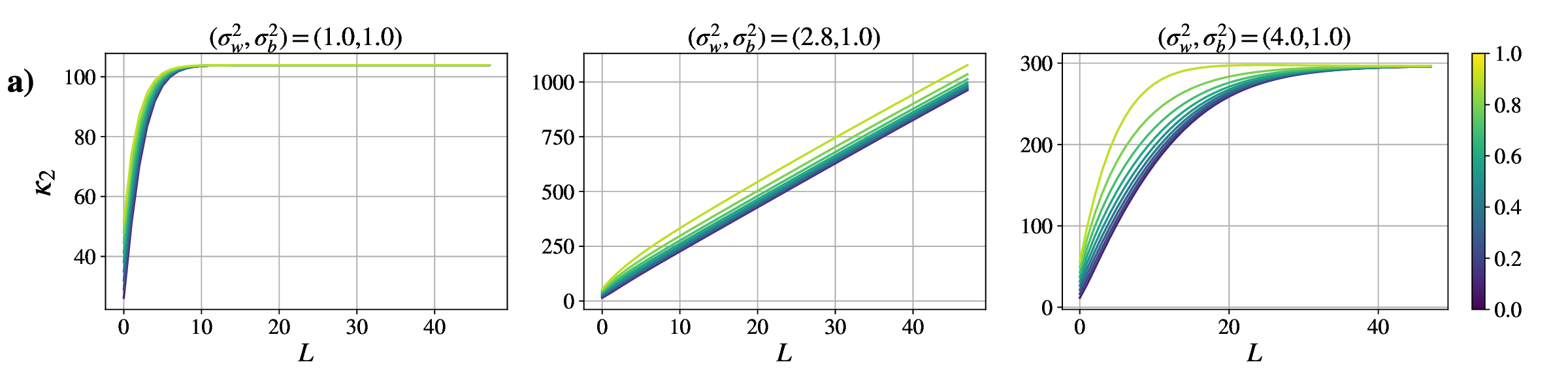}
\includegraphics[width=1.\linewidth]{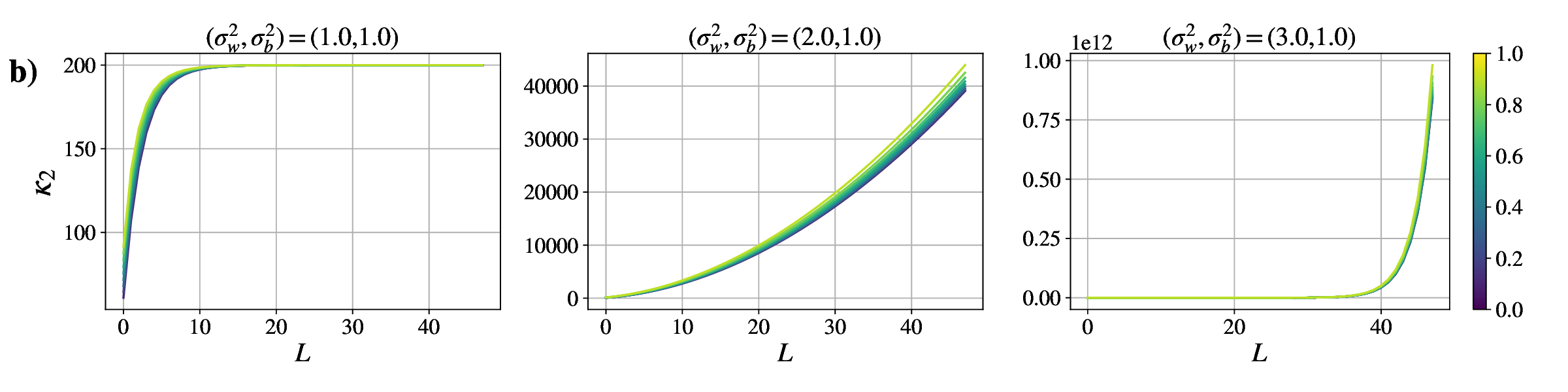}
\end{center}
  \caption{ $\kappa_2$ as a function of depth for a) \texttt{erf}, b) ReLU networks. The colorbar shows the initial value of the covariance between inputs $x_s^Tx_r \in [0,1]$. For both activation functions, $(\sigma_w^2,\sigma_b^2)$ values are chosen to lie in ordered and chaotic phases and at the border between them.}
  \label{fig:kappa_2}
\end{figure}

We can also notice that $\kappa_1$ and $q^l$ depend only on the norm of input $x$, so for normalized inputs they become data-independent. On the other hand, $\kappa_2$ and $q_{sr}^l$ depend on covariances of points in the dataset and therefore are data-dependent. However, it has also been observed in \citet{poole2016exponential} that both $q^l$ and $q^l_{sr}$ converge to their data-independent limits with depth. Let us denote their data-independent means by $\bar{q}^l$ and $\bar{q}^l_{sr}$ respectively. Then we can also write data-independent means $\bar{p}^l$ and $\bar{p}^l_{sr}$ for the backpropagated errors, as well as $\hat{\bar{q}}^l$ and $\hat{\bar{q}}_{sr}^l$ for the activations. This leads to data-independent $\bar{\kappa}_1 = \sum_{l=1}^{L} \dfrac{\alpha_{l-1}}{\alpha}\hat{\bar{q}}^{l-1}\bar{p}^l$ and $\bar{\kappa}_2=\sum_{l=1}^{L}\dfrac{\alpha_{l-1}}{\alpha}\hat{\bar{q}}^{l-1}_{sr}\bar{p}_{sr}^l$. We also notice that the changes in $\kappa_2$ that come from the changes in covariance are small with respect to its mean value $\bar{\kappa}_2$ for ReLU and \texttt{erf} networks\footnote{We expect \texttt{tanh}-networks that we studied empirically in other sections to behave similar to \texttt{erf}-networks.}. Note that for these two activation functions, we can take the integrals in (\ref{eq:q_l}), (\ref{eq:q_sr_l}), (\ref{eq:p_l}) and (\ref{eq:p_sr_l}) analytically (see Appendix \ref{appendix:integrals}) and calculate $\kappa_2$ for different values of the inputs' covariance, which is shown in Figure \ref{fig:kappa_2} for ordered and chaotic phases and at the border between them.
Therefore, we can write {the} NTK as a sum of its data-independent part and a data-dependent perturbation:
\begin{equation*}
\begin{split}
  \Theta^{*}(X) = \bar{\Theta}^{*}(\mathbf{I}_S + \epsilon (X)),\\
   \bar{\Theta}^{*} = \alpha M \bigl( (\bar{\kappa}_1-\bar{\kappa}_2)\mathbf{I}_S + \bar{\kappa_2} \mathbbm{1}_S\mathbbm{1}_S^T\bigr).
\end{split}
\end{equation*}
 We note that this result about the structure of {the} NTK is consistent with the analysis of \citet{xiao2019disentangling}, where the authors study {the} NTK at large depths. 
 
 \begin{figure}
\begin{center}
\includegraphics[width=1.\linewidth]{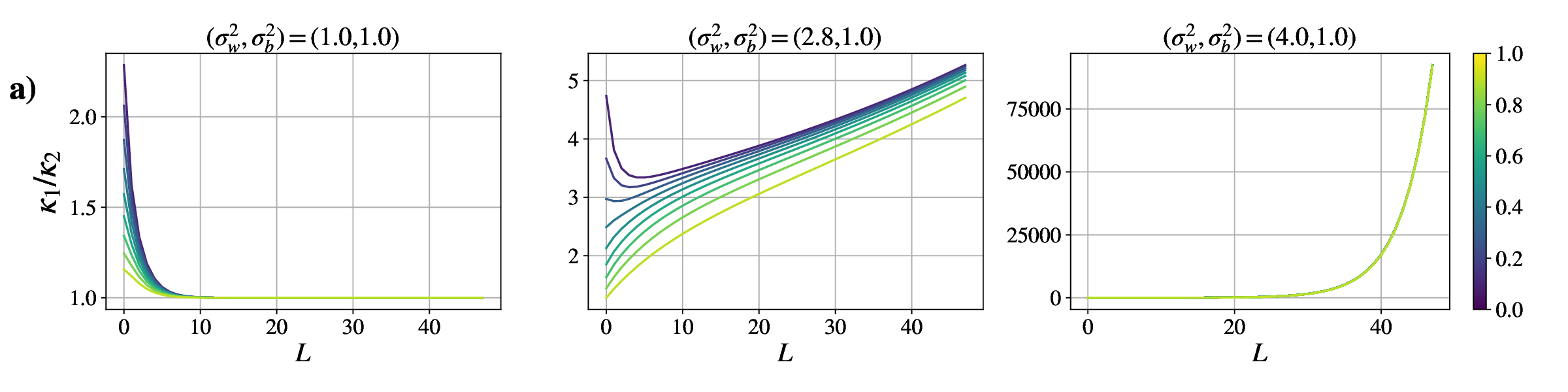}
\includegraphics[width=1.\linewidth]{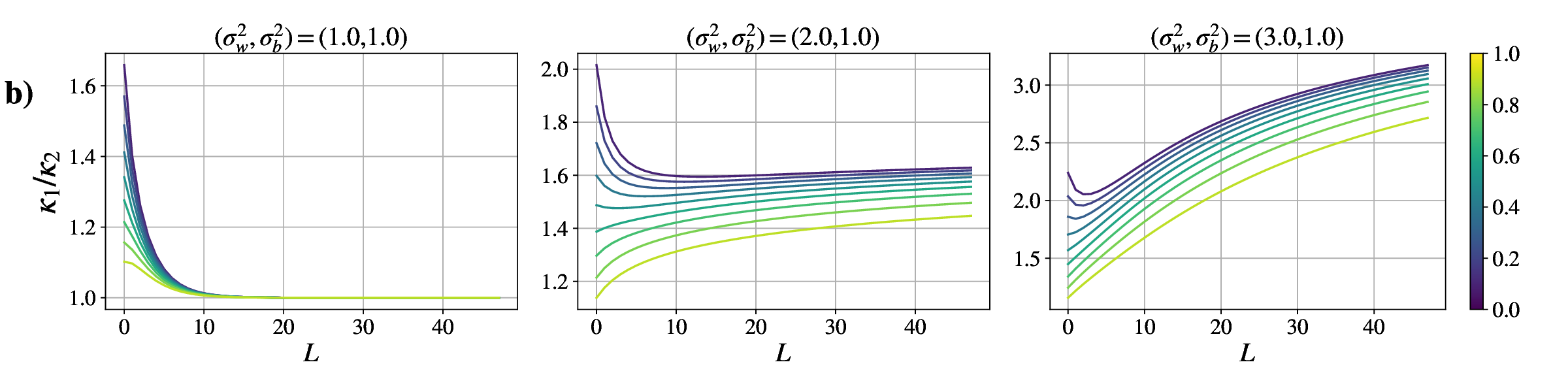}
\end{center}
  \caption{ $\kappa_1/\kappa_2$ ration as a function of depth for a) \texttt{erf}, b) ReLU networks. The colorbar shows the initial value of the covariance between inputs $x_s^Tx_r \in [0,1]$. For both activation functions, $(\sigma_w^2,\sigma_b^2)$ values are chosen to lie in ordered and chaotic phases and at the border between them.}
  \label{fig:kappa_ratio}
\end{figure}
 
 From the structure of $\Theta^{*}$, one can see that its condition number depends on the ratio $\kappa_1/\kappa_2$: when its value is high, the NTK matrix is well-conditioned, and when the ratio approaches 1 the matrix becomes close to degenerate. Figure \ref{fig:kappa_ratio} shows $\kappa_1/\kappa_2$ ratio as a function of depth for \texttt{erf} and ReLU networks in ordered and chaotic phases and at the border between them. One can see from the graphs that {the} NTK matrix is well-conditioned in the chaotic phase and ill-conditioned in the ordered phase. Ill-conditioned NTK also implies that the maximum learning rate which allows to train the network is small \citep{xiao2019disentangling,karakida2018universal}. Therefore networks in the ordered phase rapidly become untrainable with depth, which is consistent with our observations in Section \ref{section:the NTK_change_train}.

 The following theorem characterizes the dependence of the variance of the output function $f^{t=\infty}(x)$ on the data-independent part of the NTK.
\begin{theorem}\label{theorem:var_f}
Suppose a network evolves according to NTK theory under gradient flow and is fully trained ($t\to\infty$) on a dataset of size $S$. Suppose also that the NTK matrix is well-conditioned. Then the variance of its output is characterized by: 
\begin{equation*}
\begin{split}
  Var(f^{t=\infty}(x))\approx (1+\frac{A^2}{S})(\bar{q}^L - \bar{q}^L_{sr}) + (A-1)^2 \bar{q}_{sr}^L,
\end{split}
\end{equation*}
where $A = A(\kappa_1,\kappa_2) = \frac{S}{\bar{\kappa}_1/\bar{\kappa}_2+(S-1)}$.
\end{theorem}
We give a proof for this result in the {A}ppendix \ref{appendix:var_f}. 
{In the next paragraphs,} we analyze the behavior of the given variance expression and the applicability of the theorem in different situations:
\begin{itemize}
  \item \textbf{Ordered phase.} One can notice that in the ordered phase $A(\kappa_1,\kappa_2)$ converges to $1$ rapidly with depth, as $\bar{\kappa}_1/{\bar{\kappa}_2} \to 1$. This implies $Var(f^{t=\infty}(x))\propto \bar{q}^L - \bar{q}^L_{sr}$, i.e. the variance is small and decreases with depth. However, {the} NTK is also ill-conditioned, therefore small data-dependent changes can cause significant changes in the output function. Thus, the data-independent estimate for variance given by NTK theory does not explain the behavior of DNNs in the ordered phase and it is important to take into account data-dependent effects. 
  
  \item \textbf{Chaotic phase.} In the chaotic phase, {the} NTK is well-conditioned for any depth. However, only networks with depth to width ratio $L/M \approx 0$ behave as NTK theory suggests under gradient flow in the chaotic phase according to our experiments. As we saw in the previous sections, {the} NTK changes significantly during training and is random at initialization for deep networks, therefore the expression for the output function after training (\ref{eq:f_t}) does not hold. The ratio $\bar{\kappa}_1/\bar{\kappa}_2$ increases with depth in the chaotic phase, so $A(\kappa_1,\kappa_2)$ decreases, and $\bar{q}^L$ is much larger than $\bar{q}_{st}^L$ \citep{poole2016exponential}. Therefore the data-independent variance $Var(f^{t=\infty}(x)) \propto \bar{q}^L$ is high and proportional to the variance of outputs of a randomly initialized network. This is consistent with observations in \citet{chizat2019lazy} and \cite{xiao2019disentangling}. Thus, NTK theory can explain poor generalization, which shallow wide networks in the chaotic phase display. However, deeper networks may have very different behavior due to randomness at initialization and changes during gradient descent training, so they require more investigation.
  \item \textbf{EOC.} At EOC, the conditioning of the NTK at as a function of depth is similar to the chaotic phase: $\bar{\kappa}_1/\bar{\kappa}_2$ grows with depth, hence the kernel is well-conditioned. However, at EOC $\bar{q}^L$ is smaller than in the chaotic phase \citep{poole2016exponential}. This implies that networks initialized close to EOC generalize better than networks in the chaotic phase and at the same time remain trainable at large depths. We observed in the previous sections that at the border between phases NTK theory gives an approximation of network's average behavior even for deep networks, but the finite-width effects can still be significant and should be considered. 
\end{itemize}

\section{Conclusions and future work}
 In this work, we have shown that NTK theory does not generally describe the training dynamics of finite-width DNNs accurately. Only relatively shallow networks and deep networks in the ordered
phase, i.e. initialized with small $\sigma_w^2$, behave as NTK theory suggests under gradient descent. The analysis of the data-independent variance of the output function based on NTK theory shows that it is proportional to the output variance at initialization $q^L$ in the chaotic phase and at EOC. This result is not surprising, in a sense that it does not explain how training effects NNs' performance. It would provide more insight into networks' behavior if we could understand the data-dependent changes in {the} NTK that are significant {for deep networks at EOC and shallow networks in the chaotic phase} and study how these changes impact the output function. To study deep networks in the chaotic phase and at EOC, it is also essential to account for randomness in the NTK matrix at initialization and its changes during training, which cannot be done within NTK theory. Thus, an entirely new conceptual viewpoint is required to provide a full theoretical analysis of DNNs behavior under gradient descent.


\bibliographystyle{plainnat}
\bibliography{references}

\appendix

\section{Lemma \ref{lemma:Theta}} \label{appendix:the NTK}
By definition, each component of the NTK matrix is a scalar product of network's gradient vectors:
\begin{equation*}
\Theta^{0}(X) = [\nabla_w f^{0}(x_s)^T\nabla_w f^{0}(x_r)]_{x_s\in X, x_r \in X}.
\end{equation*}
In Section \ref{section:mean_field_theory} we show for the network's gradients that 
\begin{equation*}
\begin{split}
  \mathbb{E}\Bigl[(\dfrac{\partial f^0(x)}{\partial \W_{ij}^l})^2 \Bigr] &= \mathbb{E}[(\delta_i^l)^2]\mathbb{E}[(\phi(\h_j^{l-1}))^2] = \frac{1}{M_l}p^l(x) \hat{q}^{l-1}(x),\\
   \mathbb{E}\Bigl[(\dfrac{\partial f^0(x)}{\partial \mathbf{b}_{i}^l})^2 \Bigr] &= \mathbb{E}[(\delta_i^l)^2] = \frac{1}{M_l}p^l(x),
\end{split}
\end{equation*}
and similarly 
\begin{equation*}
\begin{split}
  \mathbb{E}\Bigl[\dfrac{\partial f^0(x_s)}{\partial \W_{ij}^l}\dfrac{\partial f^0(x_r)}{\partial \W_{ij}^l}\Bigr] &= \mathbb{E}[\delta_i^l(x_s)\delta_i^l(x_r)]\mathbb{E}[\phi(\h_j^{l-1})(x_s)\phi(\h_j^{l-1})(x_r)] \\
  &= \frac{1}{M_l}p_{sr}^l(x_s,x_r) \hat{q}_{sr}^{l-1}(x_s,x_r), \\
  \mathbb{E}\Bigl[\dfrac{\partial f^0(x_s)}{\partial \mathbf{b}_{i}^l}\dfrac{\partial f^0(x_r)}{\partial \mathbf{b}_{i}^l}\Bigr] &= \mathbb{E}[\delta_i^l(x_s)\delta_i^l(x_r)] = \frac{1}{M_l}p_{sr}^l(x_s,x_r).
\end{split}
\end{equation*}
Thus, we get the following expression for non-diagonal elements of {the} NTK:
\begin{equation*}
\begin{split}
  \Theta^0(x_s,x_r) &= \sum_{i,j,l} \Bigl[\dfrac{\partial f^0(x_s)}{\partial \W_{ij}^l}\dfrac{\partial f^0(x_r)}{\partial \W_{ij}^l}\Bigr] + \sum_{i,l} \Bigl[\dfrac{\partial f^0(x_s)}{\partial \mathbf{b}_{i}^l}\dfrac{\partial f^0(x_r)}{\partial \mathbf{b}_{i}^l}\Bigr]\\
  &= \sum_l M_l M_{l-1} \mathbb{E}[\delta_i^l(x_s)\delta_i^l(x_r)]\mathbb{E}[\phi(\h_j^{l-1})(x_s)\phi(\h_j^{l-1})(x_r)]\\
  &\quad + \sum_l M_l\mathbb{E}[\delta_i^l(x_s)\delta_i^l(x_r)]\\
  &= \sum_l \alpha_{l-1}M p_{sr}^l(x_s,x_r) q^{l-1}_{sr}(x_s,x_r) + \sum_l p_{sr}^l(x_s,x_r) \\
  &= \alpha M \Bigl(\sum_l \frac{\alpha_{l-1}}{\alpha} p_{sr}^l(x_s,x_r) q^{l-1}_{sr}(x_s,x_r) + O(1/M)\Bigr)\\
  &= \alpha M ( \kappa_2(x_s,x_r) + O(1/M))
\end{split}
\end{equation*}
Similarly, we get the expression for diagonal elements of the NTK matrix:
\begin{equation*}
\begin{split}
  \Theta^0(x,x) = \alpha M ( \kappa_1(x) + O(1/M)),
\end{split}
\end{equation*}
which gives the statement of the lemma.

\section{Theorem \ref{theorem:var_f}} \label{appendix:var_f}
Recall the formula of the output function after training:
\begin{equation*}
\begin{split}
  f^{t=\infty}(x) = \Theta^{0}(x,X)\Theta^{0}(X)^{-1}Y + f^0(x) - \Theta^{0}(x,X)\Theta^{0}(X)^{-1}f^0(X).
\end{split}
\end{equation*}
As initialization of the network's parameters $w_0$ is centered Gaussian, the expectation of the output at initialization is equal to zero:
\begin{equation*}
\begin{split}
  \mathbb{E}_{w_0}[f^0(x)] = 0, \quad \mathbb{E}_{w_0}[f^0(X)] = \mathbf{0}_S.
\end{split}
\end{equation*}
Then if {the} NTK is deterministic at initialization we can write the expectation as follows:
\begin{equation*}
\begin{split}
  \mathbb{E}_{w_0}[f^{t=\infty}(x)] = \mathbb{E}_{w_0}[\Theta^{0}(x,X)\Theta^{0}(X)^{-1}Y] = \Theta^{*}(x,X)\Theta^{*}(X)^{-1}Y
\end{split}
\end{equation*}
because neither $Y$ nor $\Theta^{*}$ are random with respect to the initialization parameters.

To obtain the variance of output, we also need to write the expected values of all the terms of squared $f^{t=\infty}$. First, by Lemma \ref{lemma:f0_distr}:
\begin{equation*}
\begin{split}
  \mathbb{E}_{w_0}[(f^0(x))^2] &= q^L(x). \\
\end{split}
\end{equation*}
Then,
\begin{equation*}
\begin{split}
  \mathbb{E}_{w_0}[(\Theta^{0}(x,X)\Theta^{0}(X)^{-1}Y)^2]&= (\Theta^{*}(x,X)\Theta^{*}(X)^{-1}Y)^2 =\mathbb{E}_{w_0}^2[f^{t=\infty}(x)].
\end{split}
\end{equation*}  
And
\begin{equation*}
\begin{split}
  \mathbb{E}_{w_0}[(\Theta^{0}(x,X) & \Theta^{0}(X)^{-1}f^0(X))^2] \\
  &= tr(\mathbb{E}_{w_0}[f^0(X)f^0(X)^T] \Theta^{*}(X)^{-1}\Theta^{*}(x,X)^T\Theta^{*}(x,X)\Theta^{*}(X)^{-1})\\
  &= tr( K(X)\Theta^{*}(X)^{-1}\Theta^{*}(x,X)^T\Theta^{*}(x,X)\Theta^{*}(X)^{-1}),\\
\end{split}
\end{equation*}
where
\begin{equation*}
  K(X) = \begin{bmatrix}
  q^L(x_1) & q_{sr}^L(x_1,x_2) & \dots & q_{sr}^L(x_1,x_S) \\
  q_{sr}^L(x_1,x_2) & q^L(x_2) & & \hdots \\
  \dots & & & q_{sr}^L(x_1,x_{S-1})\\
  q_{sr}^L(x_1,x_{S}) & \dots & q_{sr}^L(x_1,x_{S-1}) & q^L(x_S)&
  \end{bmatrix}.
\end{equation*}
$K(X)$ is the NNGP matrix, which characterizes the Gaussian process of a randomly initialized network. Finally:
\begin{equation*}
\begin{split}
  \mathbb{E}_{w_0}[f^0(x)\Theta^{0}(x,X)\Theta^{0}(X)^{-1}f^0(X)] &= \Theta^{*}(x,X)\Theta^{*}(X)^{-1} \mathbb{E}_{w_0}[f^0(x)f^0(X)] \\
  & = \Theta^{*}(x,X)\Theta^{*}(X)^{-1}q_{sr}^L(x,X),
\end{split}
\end{equation*}
where $q_{sr}^L(x,X) = [q_{sr}^L(x,x_s)]_{s=1,\dots S}^T$. The other terms are equal to zero. Moreover, we can see that terms of variance with $Y$ cancel each other.

We now recall that $\Theta^{*}(X) = \bar{\Theta}^{*}(\mathbf{I}_S + \epsilon(X))$ and $\bar{\Theta}^{*} = \alpha M \bigl( (\bar{\kappa}_1-\bar{\kappa}_2)\mathbf{I}_S + \bar{\kappa_2} \mathbbm{1}_S\mathbbm{1}_S^T\bigr)$. Then we can invert $\bar{\Theta}^{*}$ by Woodbury identity:

\begin{equation*}
  \bar{\Theta}^{*\ -1} = \dfrac{1}{\alpha M (\bar{\kappa}_1-\bar{\kappa}_2)} \bigl(\mathbf{I}_S - \dfrac{\bar{\kappa}_2}{\bar{\kappa}_1+(S-1)\bar{\kappa}_2}\mathbbm{1}_S\mathbbm{1}^T_S\bigr)
\end{equation*}
We assumed that the NTK matrix is well-conditioned, so the change in the $\bar{\Theta}^{*\ -1}$ caused by the perturbation term is relatively small and we can write $\Theta^{*\ -1}(X) = \bar{\Theta}^{*\ -1}(\mathbf{I}_S + \tilde{\epsilon}(X))$. Then we can also approximate the above expectation as follows:

\begin{equation*}
\begin{split}
   \Theta^{*}(x,X)\Theta^{*}(X)^{-1}q_{sr}^L(x,X) &\approx \dfrac{\bar{\kappa_2}}{ (\bar{\kappa}_1-\bar{\kappa}_2)} \mathbbm{1}^T_S \bigl(\mathbf{I}_S - \dfrac{\bar{\kappa}_2}{\bar{\kappa}_1+(S-1)\bar{\kappa}_2}\mathbbm{1}_S\mathbbm{1}^T_S\bigr)q_{sr}^L(x,X)\\
   &= \dfrac{\bar{\kappa_2}}{(\bar{\kappa}_1-\bar{\kappa}_2)} \bigl(1 - \dfrac{\bar{\kappa}_2 S}{\bar{\kappa}_1+(S-1)\bar{\kappa}_2}\bigr)\mathbbm{1}^T_S q_{sr}^L(x,X)\\
   &= \dfrac{ S}{(\bar{\kappa}_1/\bar{\kappa}_2+(S-1))} \langle q^L_{sr}(x_s,x)\rangle_{s=1,\dots S},
\end{split}
\end{equation*}
\begin{equation*}
\begin{split}
   tr( K(X)\Theta^{*}(X)^{-1}&\Theta^{*}(x,X)^T\Theta^{*}(x,X)\Theta^{*}(X)^{-1}) \\
   &\approx \dfrac{\bar{\kappa}_2^2}{(\bar{\kappa}_1 - \bar{\kappa}_2)^2} (1 - \dfrac{\bar{\kappa}_2 S}{\bar{\kappa}_1+(S-1)\bar{\kappa}_2})^2tr(K(X)\mathbbm{1}_S\mathbbm{1}_S^T)\\
   &= \dfrac{S^2}{ (\bar{\kappa}_1/\bar{\kappa}_2+(S-1))^2} (\dfrac{1}{S}\langle q^L(x_s)\rangle+ (1-\dfrac{1}{S}) \langle q^L_{sr}(x_s,x_r)\rangle ).
\end{split}
\end{equation*}
Taking expectation of the above expressions over a random dataset $D$, which is independent to random initialization $w_0$, we get 

\begin{equation*}
\begin{split}
  \mathbb{E}_{w_0,D}[f^0(x)\Theta^{0}(x,X)\Theta^{0}(X)^{-1}f^0(X)]
  &= \dfrac{ S}{\bar{\kappa}_1/\bar{\kappa}_2+(S-1)} \mathbb{E}_X[\langle q^L_{sr}(x_s,x) \rangle] \\
  &= \dfrac{S}{\bar{\kappa}_1/\bar{\kappa}_2+(S-1)} \bar{q}^L_{sr},\\
  \mathbb{E}_{w_0,X}[(\Theta^{0}(x,X)\Theta^{0}(X)^{-1}f^0(X))^2]&= \dfrac{S^2}{ (\bar{\kappa}_1/\bar{\kappa}_2+(S-1))^2}\cdot\\
  &\quad\cdot \mathbb{E}_{X}(\dfrac{1}{S}\langle q^L(x_s)\rangle+ (1-\dfrac{1}{S}) \langle q^L_{sr}(x_s,x_r)\rangle )\\ 
  &= \dfrac{S^2}{ (\bar{\kappa}_1/\bar{\kappa}_2+(S-1))^2} \bigl(\frac{1}{S} \bar{q}^L + (1-\frac{1}{S})\bar{q}_{sr}^L \bigr).
\end{split}
\end{equation*}
Putting everything together, we get

\begin{equation*}
\begin{split}
  \mathbb{E}_{w_0,X}[(f^{t=\infty}_{lin}(x))^2] - \mathbb{E}_{w_0,X}[f^{t=\infty}_{lin}(x)]^2 \approx \bar{q}^L - 2 \dfrac{S}{\bar{\kappa}_1/\bar{\kappa}_2+(S-1)} \bar{q}^L_{sr}\\
  + \dfrac{S^2}{ (\bar{\kappa}_1/\bar{\kappa}_2+(S-1))^2} \bigl(\frac{1}{S} \bar{q}^L + (1-\frac{1}{S})\bar{q}_{sr}^L \bigr).
\end{split}
\end{equation*}
Denoting $A = \dfrac{S}{\bar{\kappa}_1/\bar{\kappa}_2+(S-1)}$, we can rewrite the above expression as
\begin{equation*}
\begin{split}
  Var(f^{t=\infty}(x))\approx (1+\frac{A^2}{S})(\bar{q}^L - \bar{q}^L_{sr}) + (A-1)^2 \bar{q}_{sr}^L.
\end{split}
\end{equation*}

\begin{figure}
\begin{center}
\includegraphics[width=1.\linewidth]{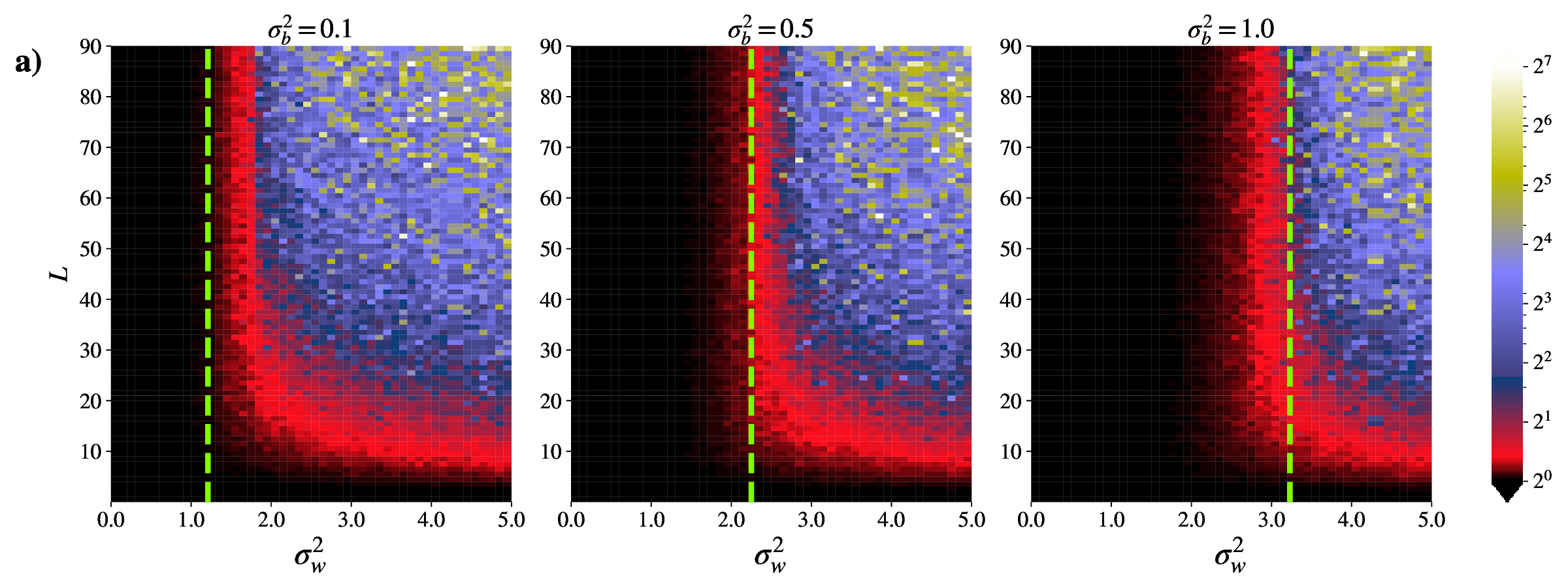}
\includegraphics[width=1.\linewidth]{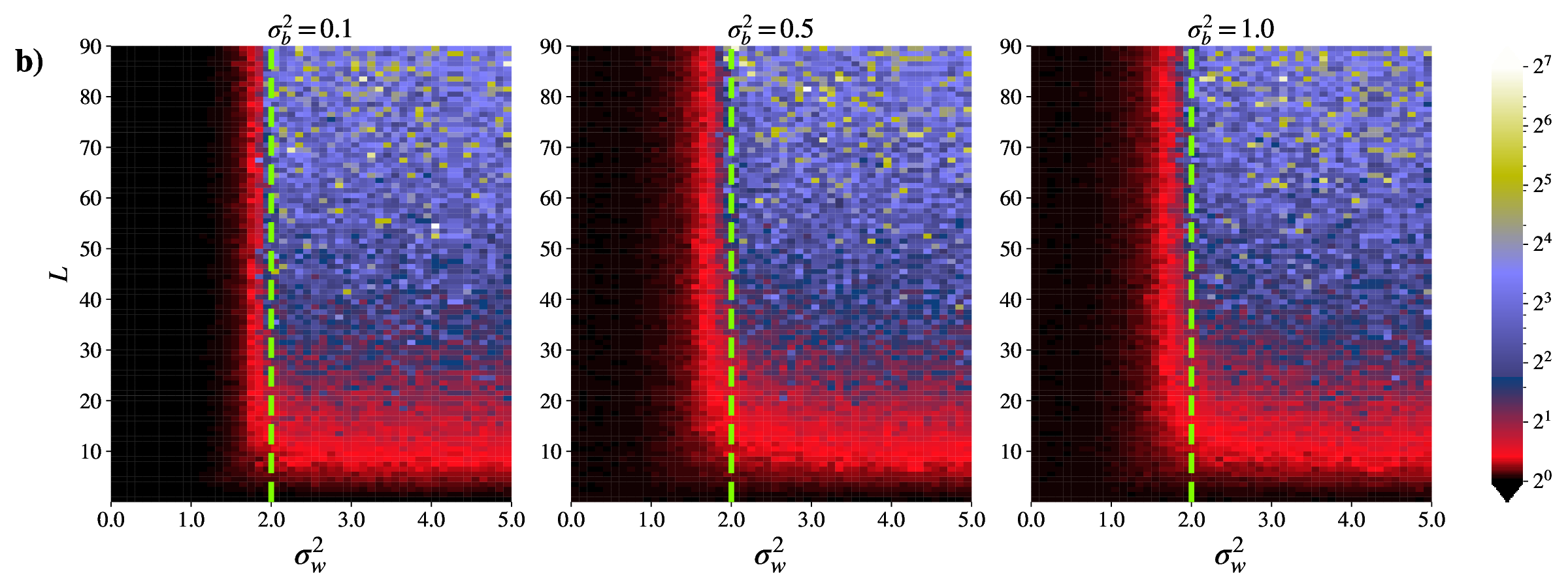}
\end{center}
  \caption{{$\dfrac{\mathbb{E}[\Theta^{0}(x,x)^2]}{\mathbb{E}^2[\Theta^{0}(x,x)]}$ ratio for fully-connected a) \texttt{tanh}, b) ReLU networks of width $M=100$ for different $\sigma_b$ values. The dashed line shows the theoretical border between ordered and chaotic phases ($\chi^l_1=1$) for the given hyperparameters. For \texttt{tanh} networks the location of the border between phases depends on $\sigma_b^2$, while for ReLU networks it is the same for all the $\sigma_b^2$ values.}}
\label{fig:b_Kxx_var}
\end{figure}

{

\section{Effects of biases initialization on the NTK variance at initialization} \label{appendix:var_b}
Figure \ref{fig:b_Kxx_var} shows the dependence of {the} NTK variance at initialization on $\sigma_b^2$. One can see that lower $\sigma_b^2$ values yield narrower boundary between the two phases, but the general picture stays similar to the one in Figure \ref{fig_relu_Kxx_var}. }

\begin{figure}
\begin{center}
\includegraphics[width=1.\linewidth]{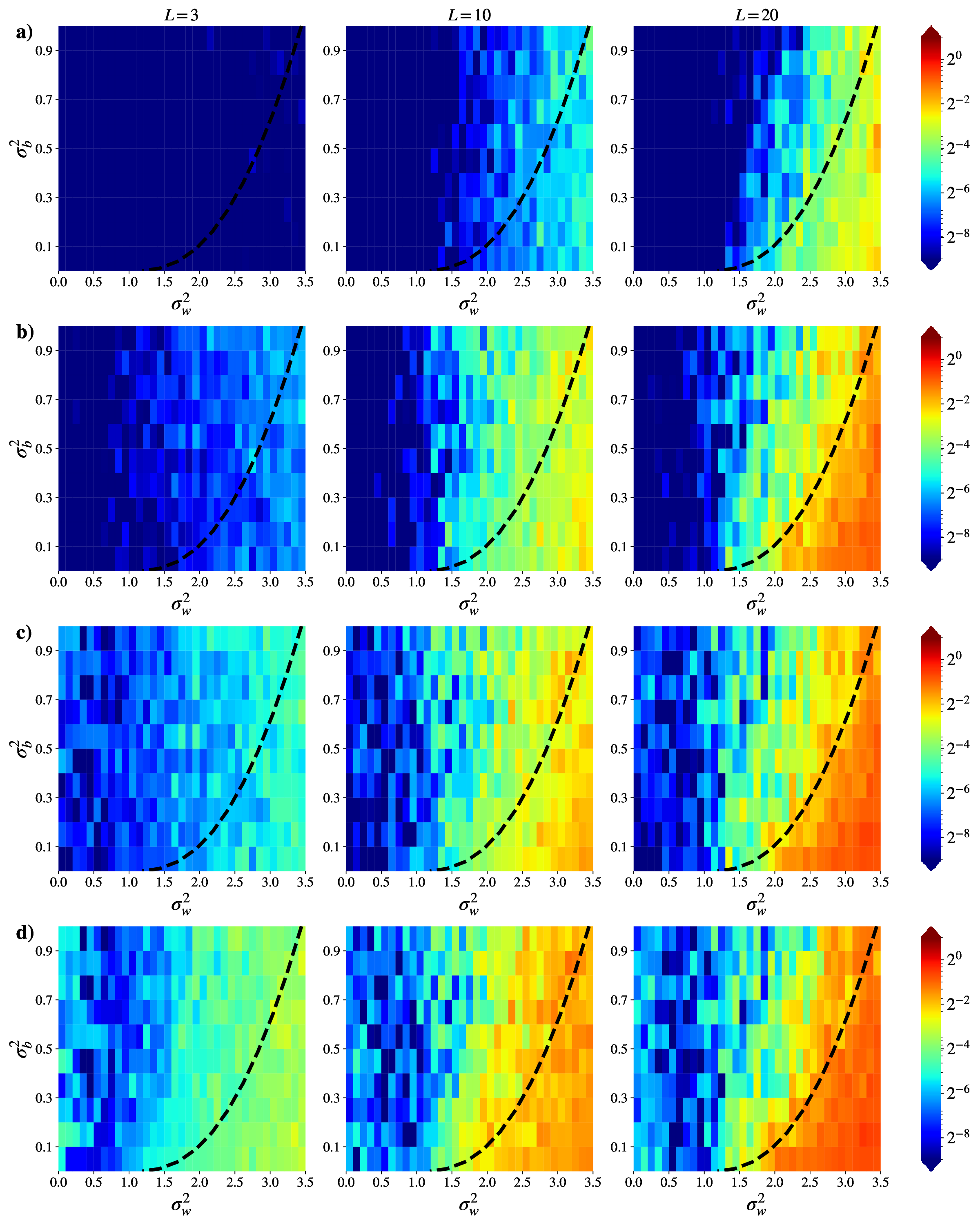}
\end{center}
  \caption{{Relative change in the NTK norm $\|\Theta^t-\Theta^{0}\|_F/\|\Theta^{0}\|_F$ for \texttt{tanh} networks after a) $10$, b) $10^2$, c) $10^3$, d) $10^4$ gradient descent steps. The training parameters are the same as in Figure \ref{fig:tanh_train_heatmap}.}}
\label{fig:tanh_train_fixedepochs}
\end{figure}

\begin{figure}
\begin{center}
\includegraphics[width=1.\linewidth]{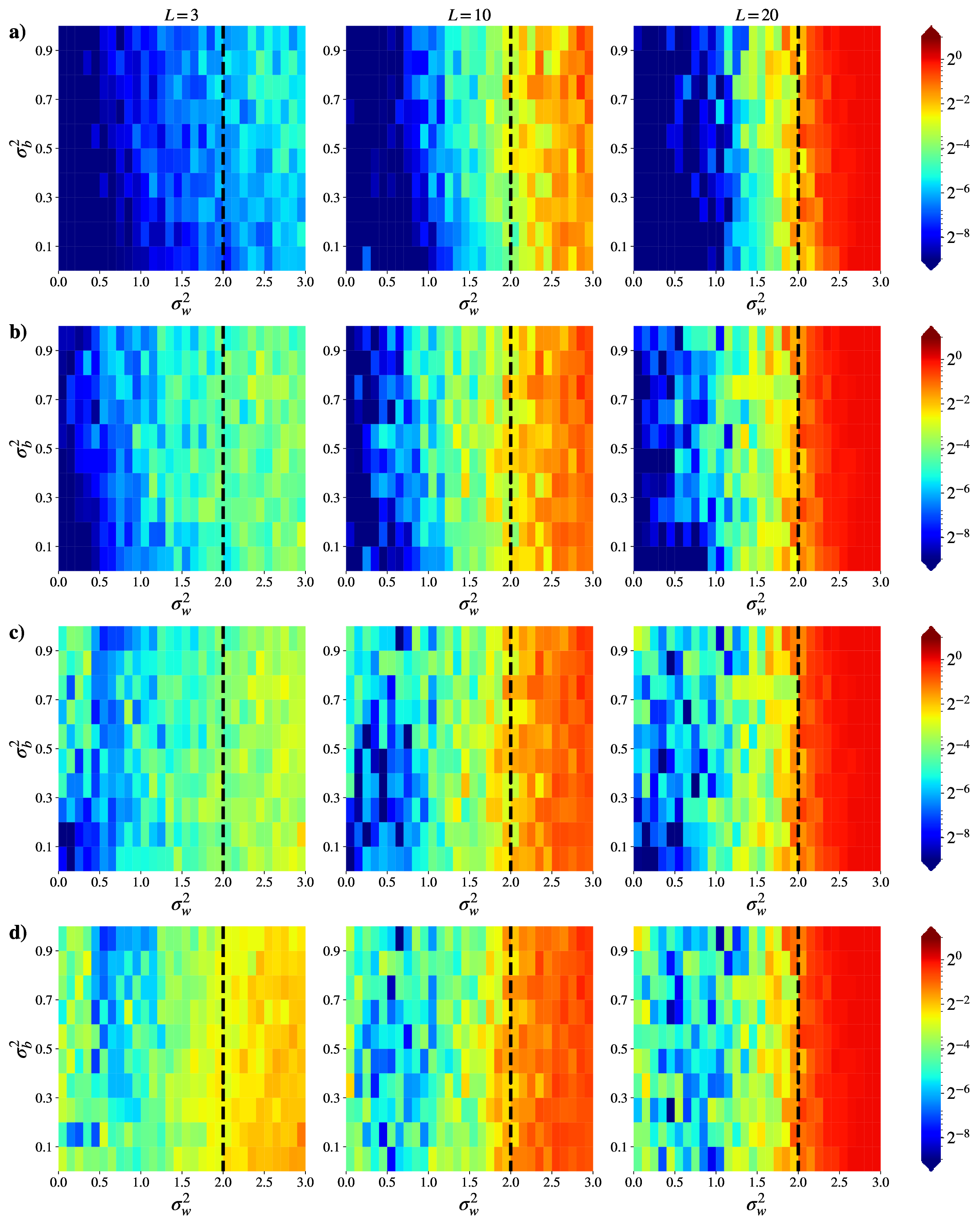}
\end{center}
  \caption{{Relative change in the NTK norm $\|\Theta^t-\Theta^{0}\|_F/\|\Theta^{0}\|_F$ for ReLU networks after a) $10$, b) $10^2$, c) $10^3$, d) $10^4$ gradient descent steps. The training parameters are the same as in Figure \ref{fig:relu_train_heatmap}.}}
\label{fig:relu_train_fixedepochs}
\end{figure}

\begin{figure}
\begin{center}
\includegraphics[width=1.\linewidth]{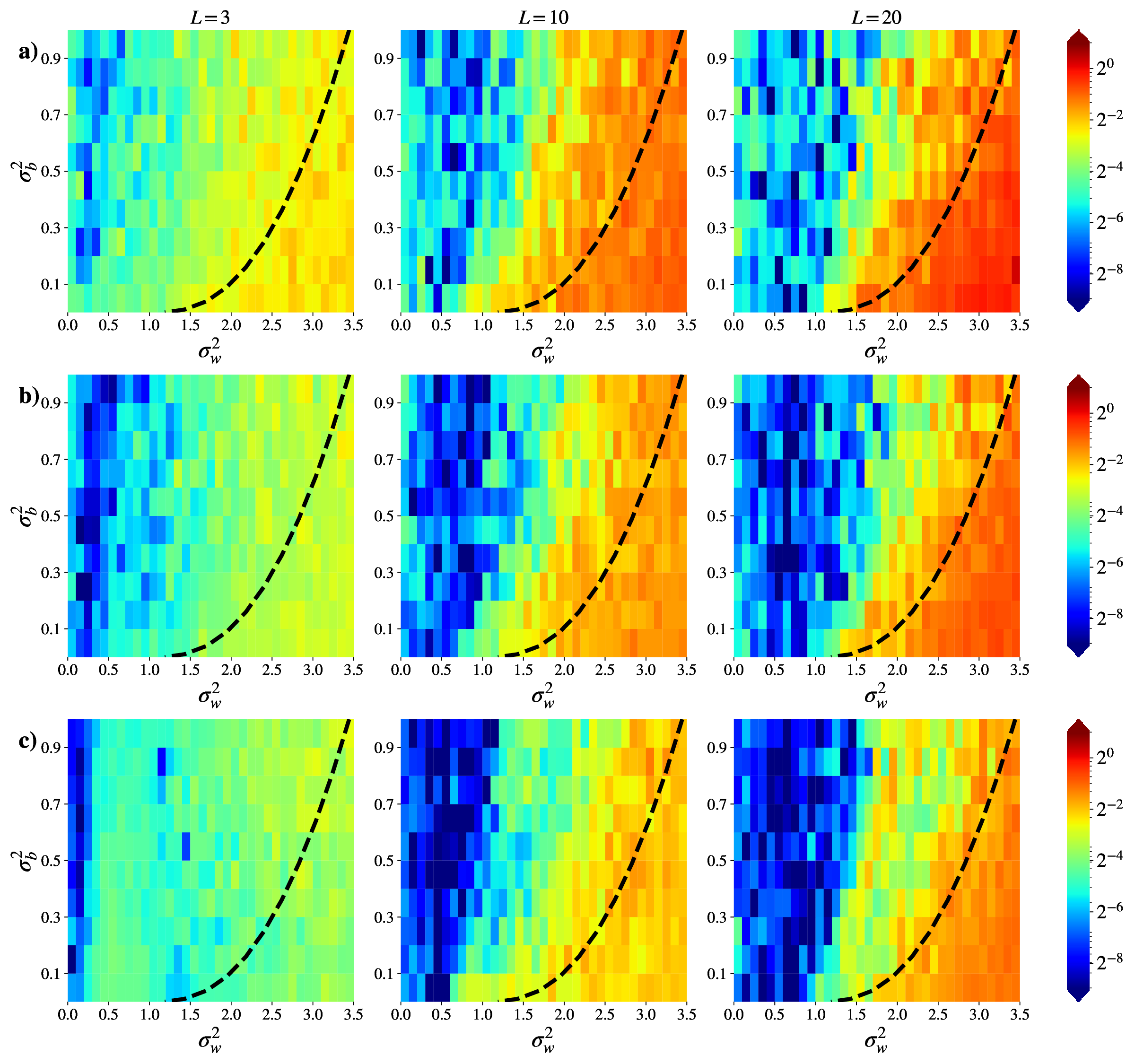}
\end{center}
  \caption{{Relative change in the NTK norm $\|\Theta^t-\Theta^{0}\|_F/\|\Theta^{0}\|_F$ for \texttt{tanh} networks of width \\ a) $M=128$, b) $M=256$, c) $M=512$ in the end of training. The training parameters are the same as in Figure \ref{fig:tanh_train_heatmap}.}}
\label{fig:tanh_Ms}
\end{figure}

\begin{figure}
\begin{center}
\includegraphics[width=1.\linewidth]{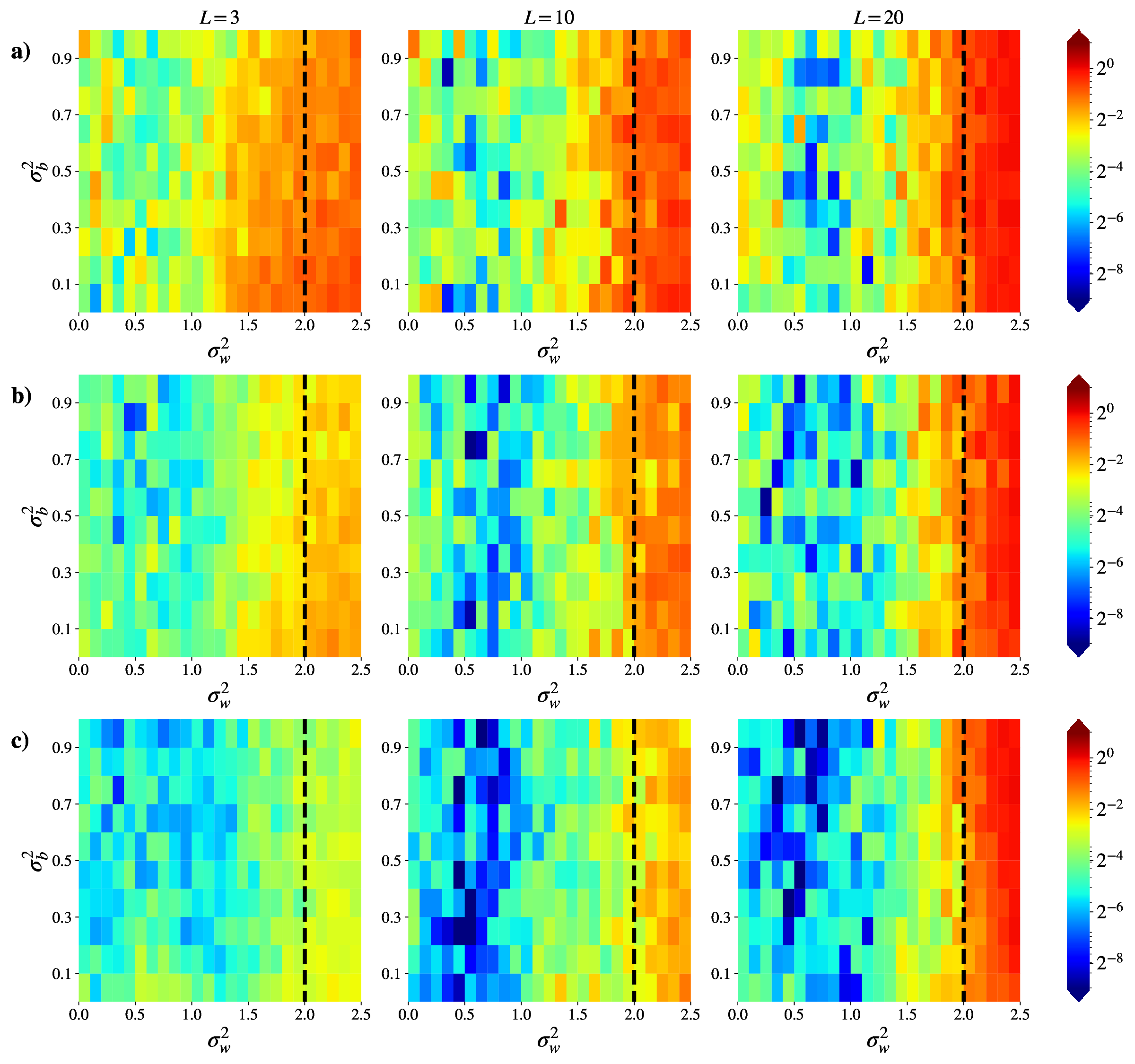}
\end{center}
  \caption{{Relative change in the NTK norm $\|\Theta^t-\Theta^{0}\|_F/\|\Theta^{0}\|_F$ for ReLU networks of width \\ a) $M=128$, b) $M=256$, c) $M=512$ in the end of training. The training parameters are the same as in Figure \ref{fig:relu_train_heatmap}.}}
\label{fig:relu_Ms}
\end{figure}

{

\section{Additional experiments on the NTK change during training}\label{appendix:the NTK_train}
Here we provide additional figures on changes of the NTK during gradient descent training.

Figures \ref{fig:tanh_train_fixedepochs} and \ref{fig:relu_train_fixedepochs} show changes in the NTK matrix norm as a function of the number of training steps for \texttt{tanh} and ReLU networks, respectively. One can see how the NTK changes after $10,10^2,10^3$ and $10^4$ training steps. The findings from these figures are similar to the analysis we provided in Section \ref{section:the NTK_change_train}: the NTK behaviour changes significantly around the border between ordered and chaotic phases. One can also see that for deep networks in the chaotic phase the NTK changes significantly already in the early stages of training, while networks in the ordered phase display very low changes in the NTK norm for a long time. 

Figures \ref{fig:tanh_Ms} and \ref{fig:relu_Ms} show the effects of the network width on the changes of the NTK matrix during training. We provide experiments for $M=128,256,512$. One can see that, as expected in NTK theory, higher $M$ values overall result in smaller changes of the NTK. However, with all the width values, one can see the transition from ordered to chaotic phase, which gets more pronounced with the network's depth.
}

\section{Analytical relations for integrals in Section \ref{section:mean_field_theory}}\label{appendix:integrals}

\subsection{ReLU networks}
ReLU activation function is defined by 
\begin{equation*}
 \phi(x) =
  \begin{cases}
   x & x > 0,\\
   0 & x \leq 0.
  \end{cases}    
\end{equation*}
Then to obtain analytical expressions for $q^l$ and $q_{sr}^l$ we can take the following integrals, which appear in (\ref{eq:q_l}) and (\ref{eq:p_l}):
\begin{equation*}
\begin{split}
  &\int Dz \cdot \phi(az)^2 = a^2/2,\\
  &\int Dz \cdot [\phi^{'}(az)]^2 = 1/2,\\
\end{split}
\end{equation*}
Then we immediately get 
\begin{equation*}
\begin{split}
  q^l &= \frac{\sigma_w^2}{2} q^{l-1} + \sigma_b^2,\\
  p^{l-1} &= \frac{\sigma_w^2}{2} p^{l} \frac{M_{l}}{M_{l+1}}.
\end{split}
\end{equation*}
 Similarly, to get analytical expressions for $q_{sr}^l$ and $p_{sr}^l$, we can take the integrals in (\ref{eq:q_sr_l}) and (\ref{eq:p_sr_l}):
\begin{equation*}
\begin{split}
  \int Dz_1 Dz_2 \cdot \phi(az_1)\phi(bz_1 + \sqrt{a^2-b^2}z_2) &= \frac{a}{2\pi}(\sqrt{1 - c^2} + c \pi/2 + c \arcsin(c)),\\
  \int Dz_1 Dz_2 \cdot \phi^{'}(az_1)\phi^{'}(bz_1 + \sqrt{a^2-b^2}z_2) &= \frac{1}{2\pi}(\pi/2 + \arcsin(c)),
\end{split}
\end{equation*}
where $c = b/a$, to obtain the following expressions:
\begin{equation*}
\begin{split}
  q_{sr}^l &= \frac{\sigma_w^2}{2\pi} q^{l-1} (\sqrt{1 - c^2}+ c\pi/2 + c\arcsin c) + \sigma_b^2,\\
  p_{sr}^{l-1} &= \frac{\sigma_w^2}{2\pi} p^{l} \frac{M_{l}}{M_{l+1}}(\pi/2 + \arcsin c), 
\end{split}
\end{equation*}
where $c = q_{st}^{l-1}/q^{l-1}$. 

Then, to compute the values of $q^l,q_{st}^l,p^l$ and $p_{st}^l$ in all the layers, we only need to set the following initial conditions: $q^0 = 1$ when data is normalized, $q_{st}^0\in [0,1]$ is the covariance between two inputs, $p^L = p_{st}^L = 1$ as the output depends linearly on the activations in the last layer.

\subsection{\texttt{Erf} networks}
Error function, which is a kind of sigmoid functions, is defined by
\begin{equation*}
 \phi(x) = \frac{2}{\sqrt{\pi}}\int_0^x e^{-t^2}dt.
\end{equation*}
Then, same as for ReLU activation, we analytically take the integrals from (\ref{eq:q_l}) and (\ref{eq:p_l}):
\begin{equation*}
\begin{split}
  &\int Dz \cdot \phi(az)^2 = \frac{2}{\pi}\arctan \frac{a^2}{\sqrt{a^2+1/4}},\\
  &\int Dz \cdot [\phi^{'}(az)]^2 = \frac{2}{\pi}\frac{1}{\sqrt{a^2+1/4}}\\
\end{split}
\end{equation*}
to obtain expressions for $q^l$ and $p^l$:
\begin{equation*}
\begin{split}
  q^l &= \frac{2\sigma_w^2}{\pi}\arctan \frac{q^{l-1}}{\sqrt{q^{l-1}+1/4}} + \sigma_b^2,\\
  p^{l-1} &= \frac{2\sigma_w^2}{\pi}p^{l}\frac{1}{\sqrt{q^{l-1}+1/4}} \frac{M_{l}}{M_{l+1}}.
\end{split}
\end{equation*}
And similarly we take the integrals in (\ref{eq:q_sr_l}) and (\ref{eq:p_sr_l}):
\begin{equation*}
\begin{split}
  \int Dz_1 Dz_2 \cdot \phi(az_1)\phi(bz_1 + \sqrt{a^2-b^2}z_2) &= \frac{2}{\pi}\arctan \frac{2 b}{\sqrt{(1+2a)^2 - 4b^2}},\\
  \int Dz_1 Dz_2 \cdot \phi^{'}(az_1)\phi^{'}(bz_1 + \sqrt{a^2-b^2}z_2) &= \frac{4}{\pi} \frac{1}{\sqrt{(1+2a)^2 - 4b^2}},
\end{split}
\end{equation*}
to obtain the analytical expressions for $q_{sr}^l$ and $p_{sr}^l$:
\begin{equation*}
\begin{split}
  q_{sr}^l &= \frac{2\sigma_w^2}{\pi} \arctan \frac{2\sqrt{q_{sr}^{l-1}}}{\sqrt{(1+2\sqrt{q^{l-1}})^2 - 4q_{sr}^{l-1}}} + \sigma_b^2,\\
  p_{sr}^{l-1} &= \frac{4\sigma_w^2}{\pi} p^{l} \frac{M_{l}}{M_{l+1}} \frac{1}{\sqrt{(1+2\sqrt{q^{l-1}})^2 - 4q_{sr}^{l-1}}}.
\end{split}
\end{equation*}
And the initial conditions can be specified in the same way as for the ReLU networks in the previous subsection.

\end{document}